\documentclass{article}

\usepackage[numbers,square]{natbib}
\usepackage[final,nonatbib]{neurips_2021_new}

\usepackage[utf8]{inputenc} 
\usepackage[T1]{fontenc}    
\usepackage{hyperref}       
\usepackage{url}            
\usepackage{booktabs}       
\usepackage{amsfonts}       
\usepackage{nicefrac}       
\usepackage{microtype,wrapfig}      

\usepackage[nonatbib]{neurips_2021_new}

\usepackage{microtype,bbm}

\usepackage[]{algorithm2e}
\usepackage[]{algorithmic}
\usepackage{graphicx}
\usepackage{booktabs}
\usepackage{hyperref}
\usepackage{pdfpages}
\usepackage{subcaption}
\usepackage{mwe}

\usepackage{float}
\usepackage{amsmath,amssymb,amsthm}

\usepackage{todonotes}

\DeclareMathOperator*{\argmaxA}{arg\,max}

\usepackage{pdfpages} 
\usepackage{pgffor} 
\makeatletter
\AtBeginDocument{\let\LS@rot\@undefined}
\makeatother
\def\supplementfilename{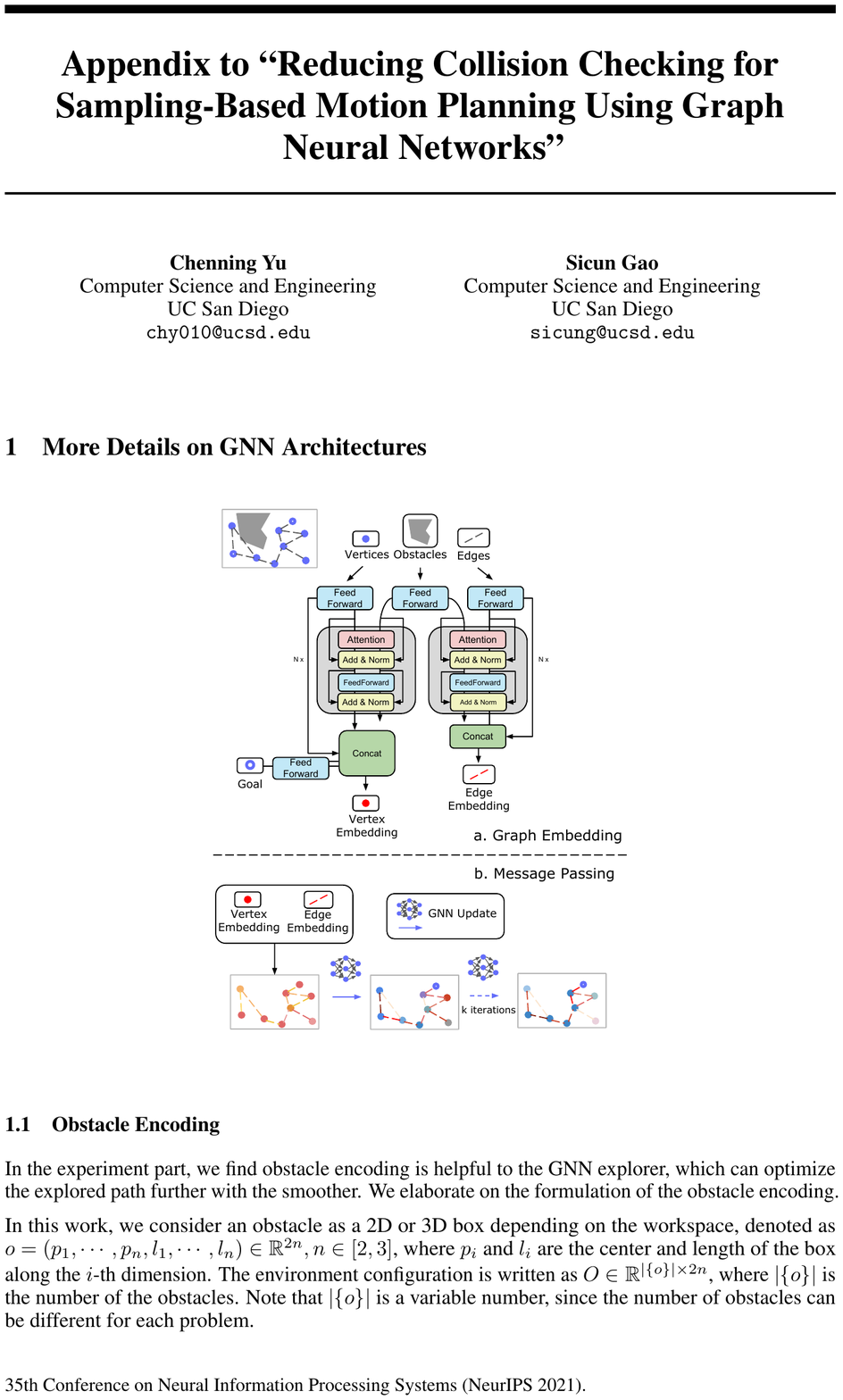}
\pdfximage{\supplementfilename}
\def\numbersupplementpages{\the\pdflastximagepages}
\newif\ifarXiv
\arXivtrue 

\title{Reducing Collision Checking for Sampling-Based Motion Planning Using Graph Neural Networks}

\author{%
  Chenning Yu \\
  Computer Science and Engineering\\
  UC San Diego\\
  \texttt{chy010@ucsd.edu} \\
  \And
  Sicun Gao\\
  Computer Science and Engineering\\
  UC San Diego\\
  \texttt{sicung@ucsd.edu} \\  
}

\begin{document}

\maketitle

\begin{abstract}
Sampling-based motion planning is a popular approach in robotics for finding paths in continuous configuration spaces. Checking collision with obstacles is the major computational bottleneck in this process. We propose new learning-based methods for reducing collision checking to accelerate motion planning by training graph neural networks (GNNs) that perform path exploration and path smoothing. Given random geometric graphs (RGGs) generated from batch sampling, the path exploration component iteratively predicts collision-free edges to prioritize their exploration. The path smoothing component then optimizes paths obtained from the exploration stage. The methods benefit from the ability of GNNs of capturing geometric patterns from RGGs through batch sampling and generalize better to unseen environments. Experimental results show that the learned components can significantly reduce collision checking and improve overall planning efficiency in challenging high-dimensional motion planning tasks. 
\end{abstract}

\section{Introduction}

Sampling-based planning 
is a popular approach to high-dimensional continuous motion planning in robotics~\cite{rrt,randomA,RRT*,DBLP:journals/corr/JansonP13,bit,ait,DBLP:books/cu/L2006}. 
The idea is to iteratively sample configurations of the robots and construct one or multiple exploration trees to probe the free space, such that the start and goal states are eventually connected by some collision-free path through the sampled states, ideally with path cost minimized. This motion planning problem is hard, theoretically PSPACE-complete~\cite{pspace-hard}, and existing algorithms are challenged when planning motions of robots with a few degrees of freedom~\cite{mani1,kino2,nonholonomic1,nonholonomic2,kino1,kino2}. In particular, the planning algorithms need to repeatedly check whether an edge connecting two sample states is feasible, i.e., that no state along the edge collides with any obstacle. This {\em collision checking} operation is the major computational bottleneck in the planning process and by itself NP-hard in general~\cite{CollisionBook,collisionNP}. For instance, consider the 7D Kuka arm planning problem in the environment shown in Figure 1. The leading sampling-based planning algorithm BIT*~\cite{bit} spends about 28.6 seconds to find a complete motion plan for the robot, in which 20.2s (70.6\% of time) is spent on collision checking. In comparison, it only takes 0.06s (0.2\% of time) for sampling all the probing states needed for constructing random graphs for completing the search. 

\begin{wrapfigure}{r}{0.45\textwidth}
\includegraphics[width=0.45\textwidth]{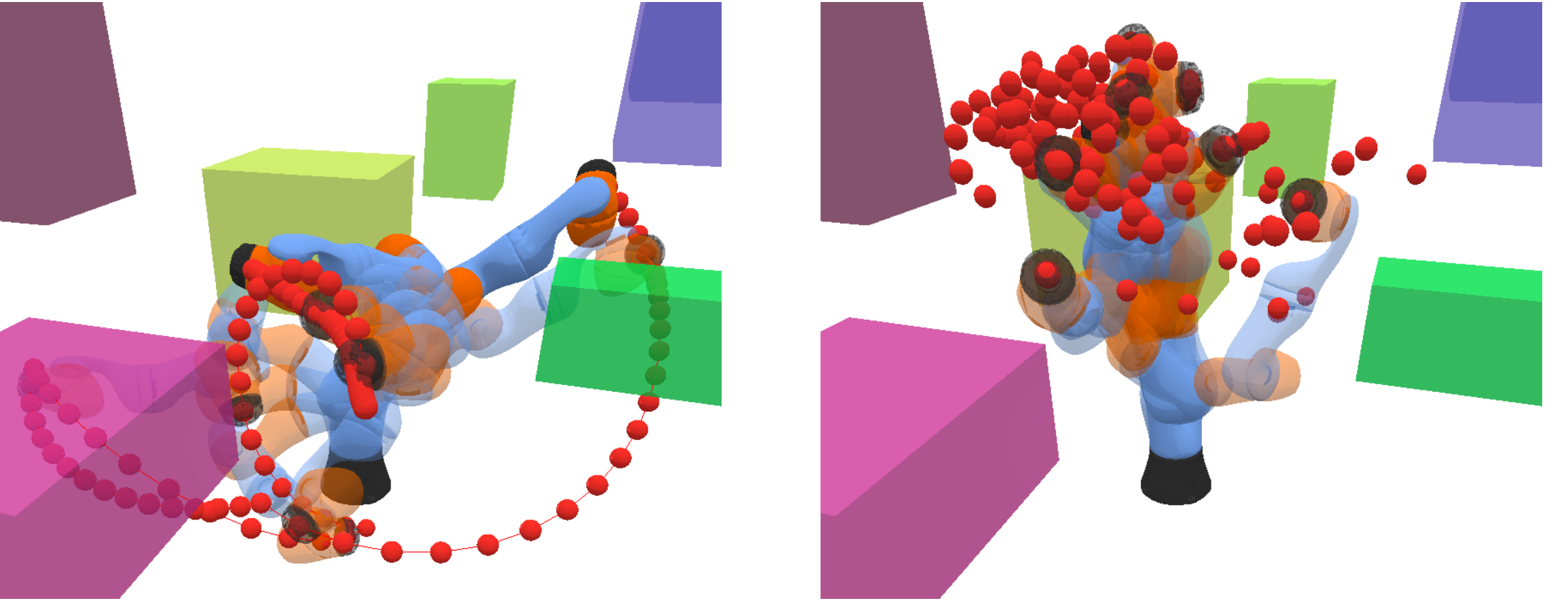}
\caption{Performance on 7D Kuka arm. Left: Trajectory generated by the proposed GNN-based approach. Right: NEXT getting stuck at a local region. Both methods were trained on the same training set.}\label{intropic}
\vspace{-10pt}
\end{wrapfigure}

Learning-based approaches have become popular for accelerating motion planning. Many recent approaches learn patterns of the configuration spaces to improve the sampling of the probing states, typically through reinforcement learning or imitation learning~\cite{RSS,CVAE,Implicit,MPNet,NEXT}. 
For instance, Ichter et~al.~\cite{CVAE} and motion planning networks \cite{MPNet} apply imitation learning on collected demonstrations to bias the sampling process. The NEXT algorithm \cite{NEXT} provides a state-of-the-art design for embedding high-dimensional continuous state spaces into low-dimensional representations, while balancing exploration and exploitation in the sampling process. It has demonstrated clear benefits of using learning-based components to reduce samples and accelerate planning. However, we believe two aspects in NEXT can be improved, if we shift the focus from reducing {\em sample complexity} to reducing {\em collision checking}. First, instead of taking the grid-based encoding of the entire workspace as input, we can use the graphs formed by batches of samples from the free space to better capture the geometric patterns of an environment. Second, having access to the entire graph formed by samples allows us to better prioritize edge exploration and collision checking based on relatively global patterns, and avoid getting stuck in local regions. In short, with reasonably relaxed budget of samples taken uniformly from the space, we can better exploit global patterns to reduce the more expensive collision checking operations instead. Figure~\ref{intropic} shows a typical example of how the trade-off benefits overall planning, and more thorough comparisons are provided in the experimental results section. 

We design two novel learning-based components that utilize Graph Neural Networks (GNNs) to accelerate the search for collision-free paths in batch sampling-based motion planning algorithms. The first component is the GNN Path Explorer, which is trained to find collision-free paths given the environment configuration and a random geometric graph (RGG) formed by probing samples. The second component is the GNN Path Smoother, which learns to optimize the path obtained from the explorer. In both models, we rely on the expressiveness and permutation invariance of GNNs as well as attention mechanisms to identify geometric patterns in the RGGs formed by samples, and accelerate combinatorial search.

The proposed learning-based components can accelerate batch sampling-based planning algorithms without compromising probabilistic completeness properties. The methods achieve higher success rate, much lower rate of collision checking, and accelerate the overall planning compared to state-of-the-art methods. 
We evaluate the proposed approach in a variety of planning environments from 2D mazes to 14D dual KUKA arms. Experiments show that our method significantly reduces collision checking, improves the overall planning efficiency, and scales well to high-dimensional tasks.  

The paper is organized as follows. We review related work and preliminaries in Section 2 and 3. We describe the detailed design of the GNN architectures in Section 4, followed by the training of GNN explorer and smoother in Section 5. We discuss experimental results in Section 6.

\section{Related Work}

\noindent{\bf Learning-based Motion Planning.} 
Learning-based approaches typically consider motion planning as a sequential decision-making problem, which could be naturally tackled with reinforcement learning or imitation learning. With model-based reinforcement learning, DDPG-MP~\cite{RSS} integrates the known dynamic of robots and trains a policy network. \citet{Obstacle} improves the obstacle encoding with the position and normal vectors. OracleNet~\cite{OracleNet} learns via oracle imitation and encodes the trajectory history by an LSTM~\cite{LSTM}.
Other than predicting nodes which are expected to greedily form a trajectory, various approaches have been designed to first learn to sample vertices, then utilize sampling-based motion planning for further path exploration through these samples. L2RRT ~\cite{L2RRT} first embeds high-dimensional configuration into low-dimensional representation, then performs RRT~\cite{rrt} on top of that. \citet{CVAE} uses conditional VAE to sample nodes. \citet{Implicit} learns a rejection sampling distribution. \citet{Adaptive} encodes the explored tree with an RNN. Motion Planning Networks \cite{MPNet} utilizes the dropout-based stochastic auto-encoder for biased sampling. NEXT~\cite{NEXT} projects the high-dimensional planning spaces into low-dimensional embedding discrete spaces, and further applies Gated Path Planning Networks~\cite{GPPN} to predict the samples.

Existing learning-based approaches have considered improving {\em collision detection}. Fastron \cite{Fastron} and ClearanceNet \cite{Clearance} learn function approximators as a proxy to collision detection, which is disparate from our focus on reducing the steps that are needed to the collision checker, and can be improved further potentially if combined together. Another recent line of work focuses on learning to explore edges given fixed graph. Value Iteration Networks~\cite{VIN} and Gated Path Planning Networks~\cite{GPPN} apply convolutional neural networks (CNN) on discrete maps, then predict the policy with a weighted attention sum over neighborhoods. Generalized Value Iteration Networks~\cite{GVIN} and \citet{NeuralExe} extend this approach for nontrivial graph by replacing CNN with GNN. However, the construction of such graphs requires ground-truth collision status for every edge on the graph at inference time.

It should be noted that other than sampling-based approaches, trajectory optimization~\cite{stomp,chomp,logicgeometric,trajopt,zhu2015convex}, and motion primitives~\cite{mueller2015computationally,zhou2019robust} are standard choices for more structured problems such as for autonomous cars and UAVs, while sampling-based methods are important for navigating high-dimensional cluttered spaces such as for manipulators and rescue robots.  

\noindent{\bf Graph Neural Networks for Motion Planning.} Graph neural networks are permutation invariant to the orders of nodes on graph, which become a natural choice for learning patterns in graph problems. They have been successfully applied in robotics applications such as decentralized control~\cite{decentral}. For sampling-based motion planning, \citet{GNNMP} utilizes GNN to identify critical samples. We focus on the different aspect of collision checking with given random geometric graphs, and can be combined with existing techniques without affecting probabilistic completeness. More broadly, GNNs have been used for learning patterns in general graph-structured problems, e.g. graph-based exploration~\cite{GraphExplore,RECON}, combinatorial optimization~\cite{CombGraph,CombGraphLinear,TransformerTSP}, neural algorithm execution~\cite{NeuralExe,GNNReason,GNNEngine,PointerGNN}. Other than to use GNN for high-dimensional planning, several works propose to first learn neural metrics, then build explicit graphs upon the learned metric which is used later to search the path~\cite{SPTM,SoRB,SGM,plan2vec}. While sharing similar interests, our work specifically focuses on how to reduce the collision checking steps for sampling-based motion planning.

\noindent{\bf Informed Sampling for Motion Planning.} A main focus in motion planning is on developing problem-independent heuristic functions for prioritizing the samples or edges to explore. Approaches include Randomized A*~\cite{randomA}, Fast Marching Trees (FMT*)~\cite{DBLP:journals/corr/JansonP13}, Sampling-based A* (SBA*)~\cite{DBLP:journals/ijrr/PerssonS14}, Batch Informed Trees (BIT*)~\cite{bit}. These methods are orthogonal to our learning-based approach, which can further exploit the problem distribution and recognize patterns through offline training to improve efficiency. Recent work in motion planning has made significant progress in reducing collision checking through batch sampling and incremental search, such as in BIT*~\cite{bit} and AIT*~\cite{ait}. The idea is to start with batches of probing random samples in the free space, and focus on reducing collision checking to edges that are likely on good paths to the goal, which also inspires our work. 

\noindent{\bf Lazy Motion Planning.} Lazy motion planning also focuses on reducing collision checking, typically with hand-crafted heuristics or features. LazyPRM and LazySP ~\cite{LazyPRM, LazySP} construct an RGG first and check the edge lazily only if that edge is on the global shortest path to the goal. Instead of calculating a complete shortest path, LWA* uses one-step lookahead to prioritize certain edges~\cite{LWA}. LRA* interleaves between LazySP and LWA*, with a predefined horizon to lookahead~\cite{LRA}. These approaches use hand-crafted heuristics and do not utilize data-dependent information from specific tasks. Recently, GLS and StrOLL \cite{gls,lazy_experience} leverage experiences to learn to select the edge to check with either fixed graphs or hand-crafted features. Our GNN-based  approach proposes the use of new representations for the learning-based  components, with the goal of directly recognizing patterns using samples from the configuration space.

\section{Preliminaries}

\paragraph{Motion Planning.} We focus on motion planning in continuous spaces, where the {\em configuration space} is $C\subseteq \mathbb{R}^n$. The configuration space includes all degrees of freedom of a robot (e.g. all joints of a robotic arm) and is different from the {\em workspace} where the robot physically resides in, which is at most 3-dimensional. For planning problem on a graph $G=\langle V,E\rangle$, we denote the start vertex and goal vertex as $v_s, v_g \in C$.
A path from $v_s$ to $v_g$ is a finite set of edges $\pi=\{e_i:(v_i,v_i')\}_{i\in [0,k]}$ such that $v_0=v_s$, $v'_k=v_g$, and $v_i'=v_{i+1}$ for all $i\in[0,k-1]$.
An environment for a motion planning problem consists of a set of obstacles $C_{obs}\subseteq C$ and free space $C_{free}=C\setminus C_{obs}$. Note that $C_{obs}$ is the projection of 3D objects in the higher-dimensional configuration space, and typically has complex geometric structures that can not be efficiently represented. 
A sample state $v\in C$ in the configuration space is free if $v \in C_{free}$, i.e., it is not contained in any obstacle. An edge connecting two samples is free if $e\subseteq C_{free}$. Namely, for every point $v$ on the edge $e$, $v \in C_{free}$. A path $\pi$ is free if all its edges are free. A random geometric graph (RGG) is a r-disc or k-nearest-neighbor (k-NN) graph $G$~\cite{rdisc,k-NN}, where nodes are randomly sampled from the free space $C_{free}$. In this paper we consider the RGG as a k-NN graph. Given a random geometric graph $G$ and a pair of start and goal configuration $(v_s, v_g)$, the goal of agent is to find a free path $\pi$ from $v_s$ to $v_g$. Without loss of generality, we consider the cost of a path to be the total length over all edges in it. 

\paragraph{Graph Neural Networks and Attention.} Let $G=\langle V,E\rangle$ be a finite graph where each vertex $v_i$ is labeled by data $x_i \in \mathbb{R}^n$. A graph neural network (GNN) learns the representation $h_i$ of node $v_i$ by aggregating the information from its neighbors $\mathcal{N}(v_i)$. Given fully-connected networks $f$ and $g$, a typical GNN encodes the representation $h_i^{(k+1)}$ of the node $v_i$ after k aggregation as:
\begin{equation}\label{eq2}
\left.\begin{aligned}
c_i^{(k)} = \oplus^{(k)}(\left\{f(h_i^{(k)}, h_j^{(k)})\mid (v_i,v_j)\in E\right\}) \mbox{ and } h_i^{(k+1)} = g(h_i^{(k)}, c_i^{(k)})
\end{aligned}\right.
\end{equation}
where $h_i^{(1)}=x_i$ and $\oplus$ is some permutation-invariant aggregation function on sets, such as max, mean, or sum. 
We will also use the attention mechanism when we need to encode a varied number of obstacles as inputs. Suppose there are $n$ keys each with dimension $d_q$: $K\in \mathbb{R}^{n\times d_k}$, each key corresponding to a value $V \in \mathbb{R}^{n\times d_v}$. Given $m$ query vectors $Q \in \mathbb{R}^{m\times d_k}$, we use a typical attention function $\text{Att}(K, Q, V)$ for each query as
$\text{Att}(K, Q, V)=\text{softmax}(QK^T/\sqrt{d_k})V$ \cite{Attention}. The function is also permutation-invariant so the order of obstacles does not affect the output.

\section{GNN Architecture for Path Exploration and Smoothing}

\subsection{Overall Approach}\label{overall}
At a high level, motion planning with batch sampling typically proceeds as follows~\cite{bit}. We first sample a batch of configurations in the free space, together with the start and goal states, and form a random graph (RGG) by connecting neighbor states (such as k-NN). A tree is then built in this graph from the start towards the goal via heuristic search. The tree can only contain collision-free edges, so each connection requires collision checking. When a path from start to goal is found, it is stored as a feasible plan, which can be later updated to minimize cost. After adding all collision-free edges in the current batch in the tree, a new batch will be added and the tree is further expanded. The algorithm keeps sampling batches and expanding the tree until the computation budget is reached. It then returns the best path found so far, or failure if none has been found.  

\begin{figure*}[h!]  
\centering
\includegraphics[width=0.28\textwidth]{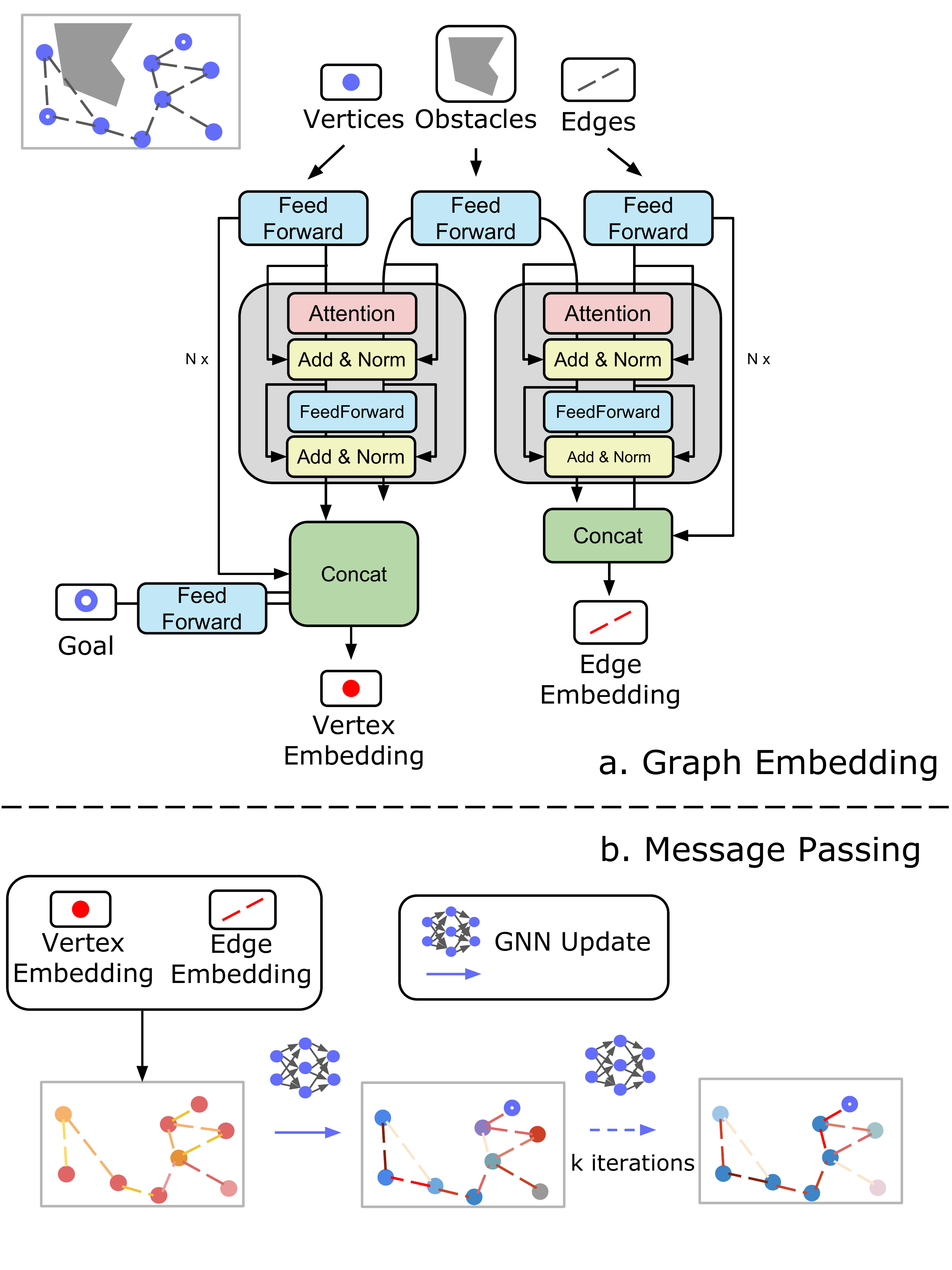}
\includegraphics[width=0.71\textwidth]{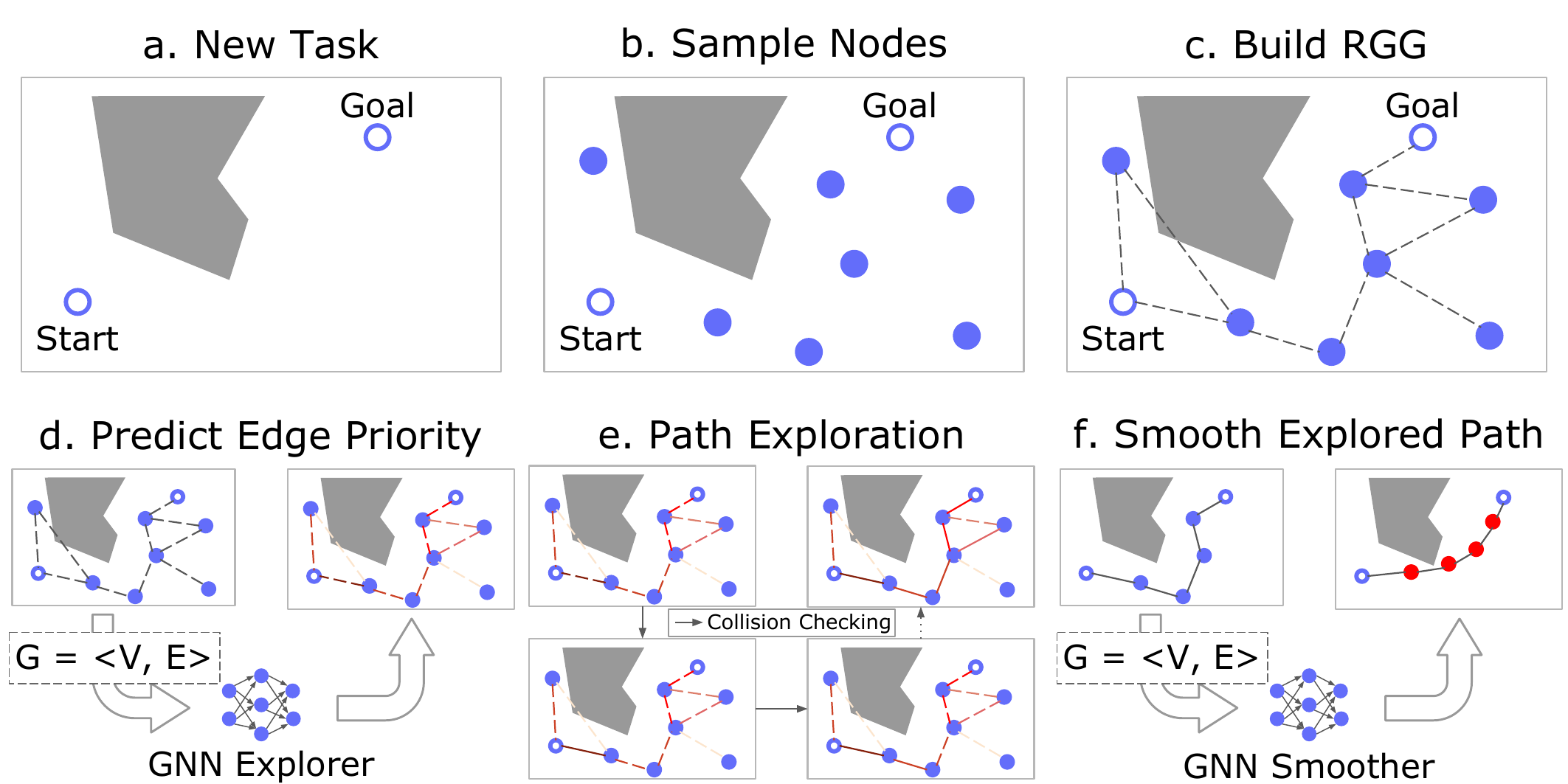}
\caption{Left: GNN architecture shared by the path explorer and the path smoother. Right (a-f): Main steps in planning with GNNs, as explained in Section~\ref{overall}.}
\label{figure:architecture}
\end{figure*}

We use GNN models to improve two important steps in this planning procedure: path exploration and path smoothing. The GNN path explorer finds a feasible path $\pi$ from start to goal given a randomly batch-sampled RGG, with the goal of reducing the number of edges that need to be checked for collision. The GNN path smoother then takes the path found by the explorer and attempts to produce another feasible path with less cost. In both tasks, the GNN models aim to recognize patterns and learn solutions to the combinatorial problems and save computation. In Figure~\ref{figure:architecture}, we illustrate the main steps for the overall algorithm. First, in (a-c), we generate an RGG composed of the vertices randomly sampled from the free space {\em without collision checking} on edges. This graph will provide patterns that the GNN explorer later uses to only prioritize certain edges for collision checking. In (d), the graph is the input to the GNN path explorer, which predicts the priority of the edges to explore and only the proposed edges are checked for collision. (e): We iteratively query the path explorer with collision checking to expand a tree in the free space until the goal vertex is reached, and sample new batches when no path is found in the current graph. (f): Once a path is provided by the path explorer, it becomes the input (together with the current RGG) to the GNN path smoother component, which outputs a new path that reduces path cost via local changes to the vertices in the input path.

\subsection{GNN Architecture}

We write the GNN path explorer as $\mathcal{N}_E$ and the GNN path smoother as $\mathcal{N}_S$. Both models take in a sampled random geometric graph $G=\langle V,E\rangle$. For an $n$-dimensional configuration space, each vertex $v_i\in \mathbb{R}^{n+3}$ contains an $n$-dimensional configuration component and a $3$-dimensional one-hot label. 
There are 3 kinds of labels for the explorer: (i) the vertices in the free space, (ii) the vertices with collision, and (iii) the special goal vertex. There are also 3 kinds of labels for the smoother: (i) the vertices on the path, (ii) the vertices in the free space, and (iii) the vertices with collision.

The vertices and the edges are first embedded into a latent space with $x\in \mathbb{R}^{|V|\times d_h}, y\in \mathbb{R}^{|E|\times d_h}$, where $d_h$ is the size of the embedding. The embeddings for the GNN explorer and smoother are different, which will be discussed later in this section. Taking the vertex and edge embedding $x, y$, the GNN aggregates the local information for each vertex from the neighbors, by performing the following operation with 2 two-layer MLPs $f_x$ and $f_y$:
\begin{equation}
\begin{split}
x_i&=g\left(x_i, \max\{f_x(x_j-x_i,x_j, x_i,y_l)\mid e_l:(v_i,v_j)\in E\}\right), \forall v_i \in V\\
y_l&= \max(y_l, f_y(x_j-x_i,x_j, x_i)), \forall e_l:(v_i,v_j)\in E
\label{coreupdate}
\end{split}
\end{equation}
Note that here we use $\max$ as the aggregation operator to gather the local geometric information, due to its empirical robustness to achieve the order invariance~\cite{PointNet}. The edge information is also incorporated by adding $y_l$ as the input to $f_x$. The update function $g$ is implemented in two different ways for the GNN explorer and smoother. Specifically, $g$ equals to the $\max$ operator for the GNN explorer, and $g(m_i, x_i)=f_g(m_i)+x_i$ as the residual connection for the GNN smoother, where $f_g$ is a two-layer MLP. We choose $\max$ operator for the explorer, due to its inductive bias to imitate the value iteration, as mentioned by~\citet{NeuralExe}. The residual connection is applied to the smoother, since intuitively the residual provides a direction for the improvement of each node on the path in the latent space, which fits our purpose to generate a shorter path for the smoother.

We also note that Equation \ref{coreupdate} directly updates on the $x$ and $y$ and is a homogeneous function similar to~\citet{HomoIterGNN}, which allows us to self-iterate $x$ and $y$ over multiple loops without introducing redundant layers. Both the GNN explorer and smoother leverage this property. After several iterations, with two MLPs $f_{\eta}$, $f_{u}$, $\mathcal{N}_E$ outputs the priority $\eta=f_{\eta}(y)$ for each edge, and $\mathcal{N}_S$ outputs a potentially shorter path $\pi'=\{u_i, u_i'\}, u_i=f_{u}(x_i)$ for $v_i \in \pi$. 

\paragraph{Special design for the GNN path explorer.} The path explorer uses the embedding of the vertices of the form $x=h_x(v, v_g, (v-v_g)^2, v-v_g)$, where $h_{x}$ is a two-layer MLP with batch normalization~\cite{BatchNorm}. Here we append the L2 distance and the difference to the goal to the vertex embedding, which serve as heuristics for the GNN to be more informed about which node is more valuable. The $y_l$ is simply computed as $y_l=h_y(v_j-v_i,v_j,v_i)$, where $h_y$ is also a two-layer MLP with batch normalization. Optionally, it is helpful for the explorer to incorporate the configuration of obstacles $O=\{\mathbf{o}\}\in \mathbb{R}^{|\{\mathbf{o}\}|\times 2n}$ as inputs, when embedding the vertices and edges. Since the obstacles of the environment has variable numbers, we utilize the attention mechanism here to update the $x$ and $y$, named as {\em obstacle encoding}, as illustrated in Figure \ref{figure:architecture}. Further details are provided in the Appendix. 

\paragraph{Special design for GNN path smoother.} The GNN smoother embeds vertices with $x=h_x(v)$, where $h_{x}$ is a two-layer MLP with batch normalization. The $y_l$ is computed as $y_l=h_y(v_j-v_i,v_j,v_i)$, where $h_y$ is a two-layer MLP with batch normalization. Each time $x$ and $y$ are updated by Equation \ref{coreupdate}, the GNN smoother will output a new smoother path $\pi'=\{(u_i, u_i')\}_{i\in[0,k]}$ , where $u_i=f_u(x_i), \forall v_i \in \pi$, given an MLP $f_u$. The $u_0$ and $u'_{k}$ are manually replaced by $v_s$ and $v_g$, to satisfy the path constraint. We assume the $\pi'$ has the same number of nodes as $\pi$. Since the GNN smoother could gain novel local geometric information with the changed vertices of the new path, we dynamically update $G=\langle V,E\rangle$, via (i) replacing those nodes labeled as path nodes in $V$ by the nodes on new path, (ii) replacing $E$ by generating a k-NN graph on the updated $V$. With the updated graph $G$, we repeat the above operation. During training, the GNN smoother outputs $\pi'$, after a random number of iterations (between 1 and 10). During evaluation, the GNN smoother outputs $\pi'$ after only one loop for each calling.

\section{Training the Path Explorer and Smoother}

Due to space limitation we provide the pseudocode for all algorithms in the Appendix. 

\subsection{GNN Explorer $\mathcal{N}_E$: Training and Inference}

The path explorer constructs a tree through sampled states with the hope of reaching the goal state in a finite number of steps. We initialize the tree $\mathcal{T}_0$ with the start state $v_s$ as its root. Every edge $e_{\mathcal{T}_i}$ in the tree $\mathcal{T}_i$ exists only if $e_{\mathcal{T}_i}$ is in the free space $C_{free}$. Given an RGG $G=\langle V, E\rangle$, Our goal is to find a tree containing the goal configuration $v_g$ by adding edges from $E$ to the tree, with as few collision checks as possible. We write the edge on frontier of the tree as $E_{f}(\mathcal{T})=\{(v_i, v_i')\mid v_i \in V_{\mathcal{T}}, v_i' \not \in V_{\mathcal{T}}\}$. We denote the set of edges with unknown collision status at time step $i$ as $E_i$. 

\paragraph{Training procedures.} Each training problem consists of a set of obstacles $O$, start vertex $v_s$, goal vertex $v_g$, we sample a k-NN graph $G=\langle V, E\rangle$, where $V$ is the random vertices sampled from the free space combined with $\{v_s, v_g\}$. The goal is to train $\mathcal{N}_E$ to predict exploration priority $\eta\in \mathbb{R}^{|E|}$. 

A straightforward way for supervision is to use the Dijkstra's algorithm to compute the shortest feasible path from $v_s$ to $v_g$, and maximize the corresponding values of $\eta$ at the edges of this path, via cross entropy loss or Bayesian ranking~\cite{BPR}. However, it does not provide useful guidance when the search tree deviates from the ideal optimal path at inference time. Instead, we first explore the graph using $\eta$ with $i$ steps, which forms a tree $\mathcal{T}_i$, where $i$ is a random number. The oracle provides the shortest feasible path $\pi_N$ in this tree and connects one of the nodes on $\mathcal{T}_i$ to the goal vertex $v_g$. We formulate this optimal path as $\pi_{N}=\{e_{N_i}:(v_{N_i},v_{N_i}')\}_{i\in[0, k]}$, where $v_{N_0} \in V_{\mathcal{T}_i}, v_{N_k}'=v_g$. We train the explorer to imitate this oracle. Namely, the explorer will directly choose $e_{N_0}\in \pi_{N}$ as the next edge to explore, among all possible edges on the frontier of $\mathcal{T}_i$, i.e. $E_i \cap E_f(\mathcal{T}_i)$. We maximize the $\eta_{N_0}$ among the values of $\{\eta_i\mid e_i \in E_i \cap E_f(\mathcal{T}_i)\}$ using the following cross entropy loss:
\begin{equation}
L_{\mathcal{N}_E} = -\log{\gamma_{N_0}} \mbox{, where }  \gamma_k = \frac{e^{\eta_k}}{\sum_{e_j \in E_i \cap E_f(\mathcal{T}_i)} e^{\eta_j}}, \forall e_k \in E_i \cap E_f(\mathcal{T}_i)
\end{equation}

\paragraph{Inference procedures.} Given the GNN $\mathcal{N}_E$, the current explored tree $\mathcal{T}_i$ at step $i$, the RGG $G=\langle V, E\rangle$ including $v_s$ and $v_g$, environment configuration $O$, GNN path explorer aims to maximize the probability of generating a feasible path by adding $e_i$ from $E_i$ to tree $\mathcal{T}_i$ as:
\begin{equation}
    e_i = \argmaxA_{e_k \in E_i \cap E_{f}(\mathcal{T}_i)} \mathcal{N}_E(\eta_k \mid V, E, O)
    \label{eta}
\end{equation}where $\eta_k$ is the output of $\mathcal{N}_E$ for the edge $e_k$. After $e_i$ is proposed by GNN using Equation \ref{eta}, we check the collision of $e_i$. If $e_i$ is not in collision with the obstacles, we add the edge $e_i$ to the tree $\mathcal{T}_i$, and remove $e_i$ from $E_i$, i.e.,  $E_{\mathcal{T}_{i+1}}=E_{\mathcal{T}_{i}}\cup\{e_i\}$, and $E_{i+1}=E_{i}\setminus\{e_i\}$. If $e_i$ is in collision with obstacles, we query the path explorer for the next proposed edge using Equation \ref{eta}, where $E_i$ is updated as $E_i=E_i\setminus \{e_i\}$. The loop terminates when we find a collision-free edge, or when $E_i\cap E_{f}(\mathcal{T}_i)=\emptyset$.
When the latter happens, 
we re-sample another batch of samples, add new samples to vertices $V$, re-construct k-NN graph $G$, re-compute $\eta$, and continue to explore the path on this new graph with the explored nodes and edges.

The exploration GNN only proposes an ordering on the candidate edges, and all possible edges may still be collision checked in the worst case. Thus, if there exists any complete path in the RGG, the algorithm always finds it. Therefore, the proposed learning-based component does not affect the probabilistic completeness of sampling-based planning algorithms~\cite{choset2005principles}.

\subsection{GNN Path Smoother $\mathcal{N}_S$: Training and Inference}

The GNN $\mathcal{N}_S$ for path smoothing takes an RGG and a path $\pi$ proposed by the explorer, and aims to produce a shorter path $\pi'$. Specifically, the input is a graph $G=\langle V,E\rangle$, where $V=V_{\pi}\cup V_{f}\cup V_{c}$, $E=E_{\pi}\cup E_{fc}$. Here, $V_{f}$ and $V_c$ are reused as the same vertices in the GNN explorer, without introducing extra sampling complexity. $E_{\pi}$ is composed of those pairs of the adjacent vertices on $\pi$, and $E_{fc}$ connects each vertex in $V_{\pi}$ to their k-nearest neighbor in $V_{f}\cup V_{c}$. Intuitively, aggregating information from $V_{f}\cup V_c$ can allow GNN to identify local regions that provide promising improvement on the current path, and avoid those that may yield potential collision.

\paragraph{Training procedures.} We train the GNN path smoother $\mathcal{N}_S$ by imitating a smoothing oracle $\mathcal{S}$ similar to the approach of gradient-informed path smoothing proposed by \citet{GradSmooth}. To prepare the training set, we iteratively perform the following two operations on each training sample path. Given a feasible path $\pi$ predicted by $\mathcal{N}_E$, 
the smoothing oracle first tries to move the nodes on the path $\pi$ with perturbation within range $\epsilon$. If the new path $\pi_M$ is feasible and has cost less than $\pi$, then $\mathcal{S}$ will continue to smooth on $\pi_M$. Otherwise, $\mathcal{S}$ will continue smoothing on $\pi$ via random perturbation. After several perturbation trials, the oracle further attempts to connect pairs of nonadjacent nodes directly by a line segment. 
If such a segment is free of collision, then the original intermediate nodes will be moved on this linear segment.  
Further details are in the Appendix. 

After $\mathcal{S}$ reaches maximum iteration, the oracle will return the smoothed path $\pi_M=\{w_i, w_i'\}_{i\in [0,k]}$. 
To generate training data, we first run our trained GNN explorer for each training problem to get the initial path $\pi$. Then the oracle path $\pi_M$ and the predicted path $\pi'$ are computed, and finally the GNN is trained via minimizing the MSE loss $L_{\mathcal{N}_S}=\frac{1}{k}\sum_{i\in [1,k]}\|\mathcal{N}_S\left( u_i \mid V, E \right)-w_i \|^2_2$. 

\begin{wrapfigure}{r}{0.4\textwidth}
\includegraphics[width=0.4\textwidth]{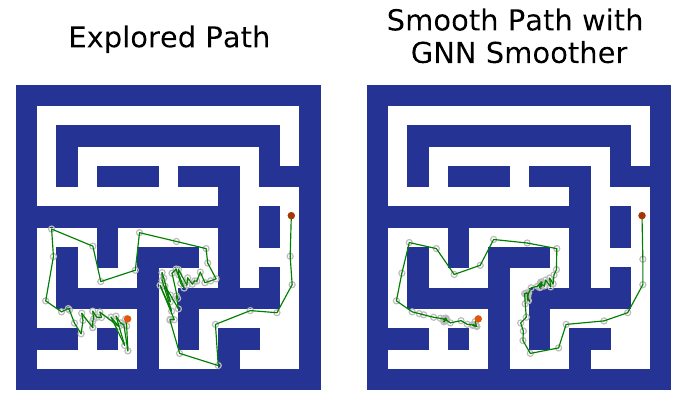}
\caption{GNN path smoother on the 2D maze problem. It learns to improve the explored path and achieves lower cost.
}\label{fig:oracle_demo}
\vspace{-10pt}
\end{wrapfigure} 
\paragraph{Inference procedures.} The GNN $\mathcal{N}_S$ for path smoothing takes an explored path $\pi:\{(v_i,v_i')\}_{i\in[0,k]}$, a sampled graph $G$, and outputs a potentially shorter path $\pi':\{(u_i,u_i')\}_{i\in[0,k]}$ which has the same number of edges as $\pi$. It is not always guaranteed that $\pi'$ is collision-free. However, such $\pi'$ still indicates directions for shortening the path, so we can improve $\pi$ towards $\pi'$ in an incremental way. The GNN smoother will try to move every node $v_i$ towards the target position $u_i$, with a small step size $\epsilon$. For each time that all the vertices are moved with a small step, we check whether each edge still holds collision-free. If not, then the vertices on the edge will undo the movement. Otherwise, the new configuration of the vertex will replace its old configuration on $\pi$. This operation will be iterated over several times, until the maximum iteration is reached, or no edges on $\pi$ can be moved further. We can then repeat the process by feeding the updated $\pi$ back to the GNN $\mathcal{N}_S$ for further improvement. The intuition here is that there might still be chances to improve upon the updated $\pi$, by aggregating new information from its changed neighborhoods. This has shown empirical advantage in our experiments. 

\begin{figure}[h!]  
\begin{center}
\centerline{\includegraphics[scale=0.32]{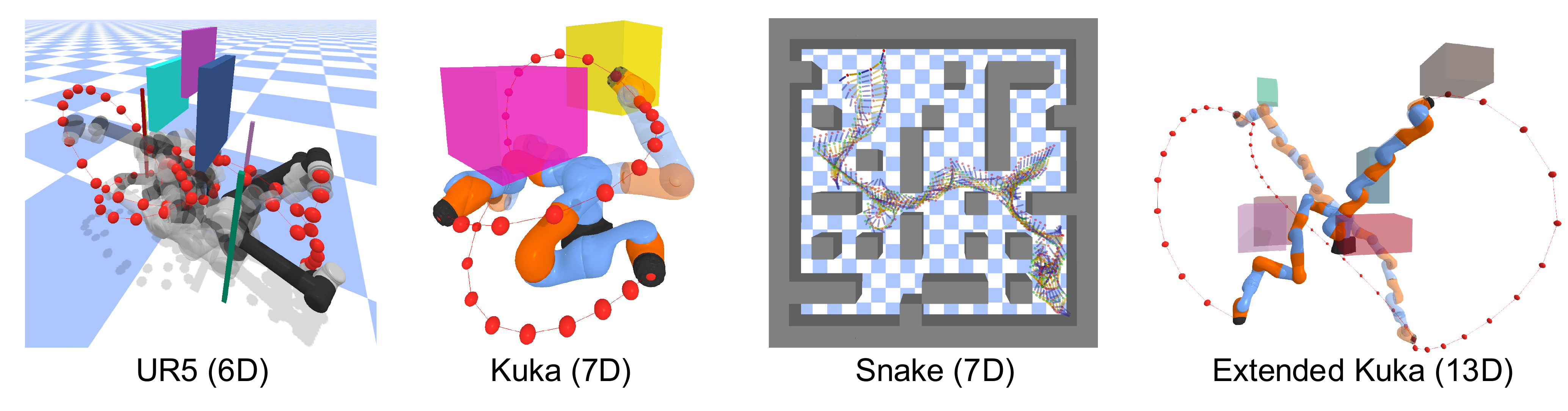}}
\label{14d}
\caption{Demonstrations for some of our environments from UR5 to 13D.
}\vspace{-0.7cm}
\end{center}
\end{figure}
\section{Experiments}

\subsection{Overall Performance}

We compare our methods with the sampling-based planning baseline RRT*~\cite{RRT*}, the state-of-the-art batch-sampling based method BIT*~\cite{bit}, the lazy motion planning method LazySP~\cite{LazySP}, and the state-of-the-art learning-based method NEXT~\cite{NEXT}. NEXT has been shown in~\cite{NEXT} to outperform competing learning-based approaches. We conduct the experiments on the following environments: (i) a 2D point-robot in 2D workspace, (ii) a 6D UR5 robot in a 3D workspace, (iii) a 7D snake robot in 2D workspace (the z-axis is fixed), (iv) a 7D KUKA arm in a 3D workspace, (v) an extended 13D KUKA arm in 3D workspace, (vi) and a pair of 7DoF KUKA arms (14 DoF) in 3D workspace.

\begin{figure*}[h]  
\centering
\centerline{\includegraphics[scale=0.766]{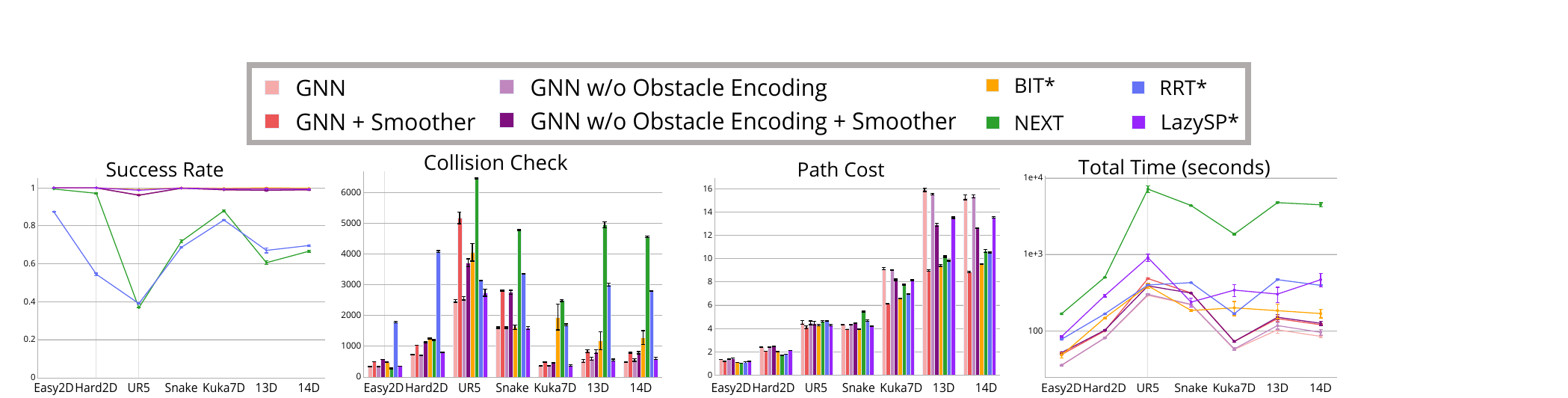}}
\caption{Comparison of performances on all environments from 2D to 14D, averaged over 4 random seeds. From left to right: (a) Success rate. (b) Collision checks. (c) Path cost. (d) Total planning time.}
\label{figure:overall}
\end{figure*}

For each environment, we randomly generate 2000 problems for training and 1000 problems for testing. Each problem contains a different set of random obstacles, and a pair of feasible $v_s$ and $v_g$.
We run all experiments over 4 random seeds. The averaged results are illustrated in Figure~\ref{figure:overall}. For the 2D environment, we directly take the training set provided by NEXT to train our GNN. We use two test sets for the 2D environment: ``Easy2D" is the original test set used for evaluating NEXT in the original paper~\cite{NEXT}, and ``Hard2D" is a new set of tests we generated by requiring the distance between the start and goal to be longer than the easy environments. The goal is to test whether the learned models can generalize to harder problems without changing the training set. Further details on the environments and hyperparameters are provided in the Appendix.

\begin{figure}[h]  
\begin{center}
\centerline{\includegraphics[scale=0.274]{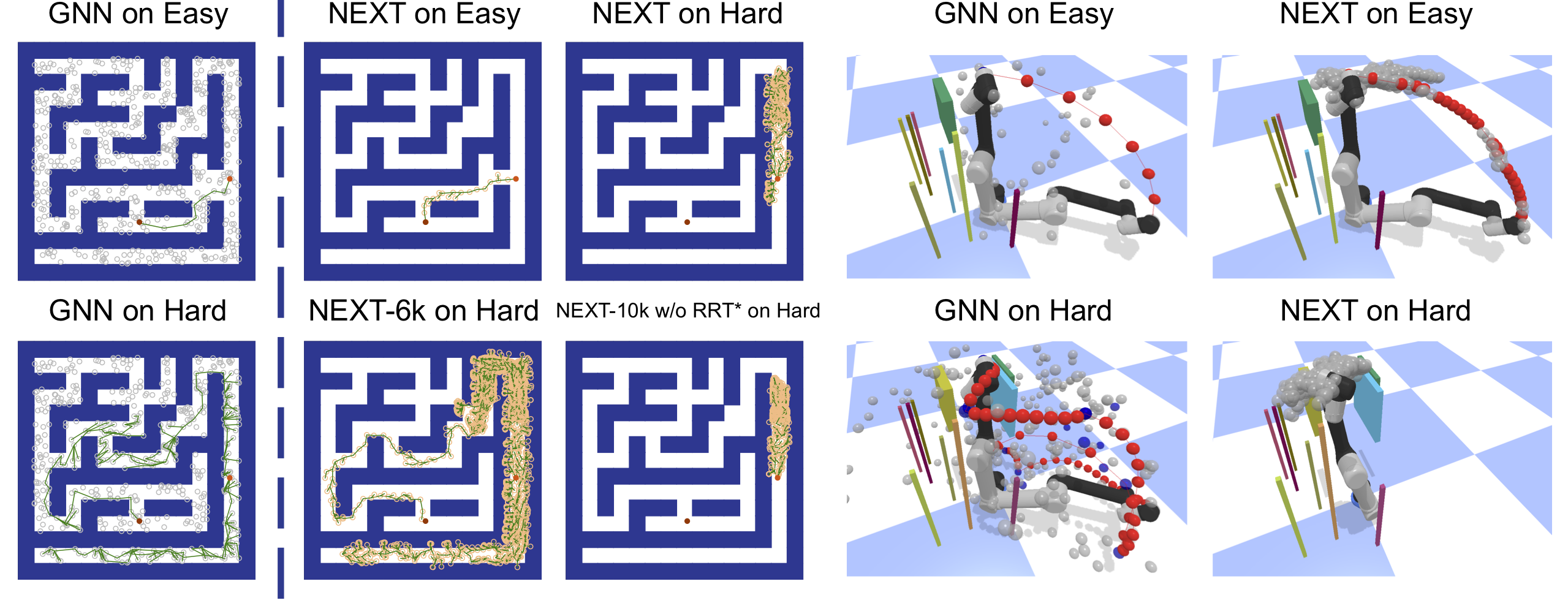}}
\caption{We test the generalization capability of GNN-based approaches and NEXT by constructing pairs of problems that have small but important difference in connectivity. The GNN models find paths on both environments quickly, while NEXT gets stuck in hard instances because of the lack of access to the graph structure provided by probing samples. The explored vertices of GNN on UR5 environment are colored in blue, and the edges on the path are colored in red.}
\label{generalization}\vspace{-0.7cm}
\end{center}
\end{figure}

\noindent{\bf Success rate.} As shown in Figure~\ref{figure:overall} (a), our method finds complete paths at 100.0\% problems on both 2D Easy and 2D Hard, and at 97.18\%, 99.85\%, 99.20\%, 99.15\%, and 99.15\% problems from UR5 to 14D, which is comparable to handcrafted heuristics used in BIT* (100.0\%, 100.0\%, 99.25\%, 99.85\%, 99.65\%, 99.92\%, 99.82\% on each environment). The learning-based planner NEXT performs well on easy 2D problems (99.37\% success rate), but drops slightly to 97.10\% on harder 2D problems, and 36.80\%, 71.80\%, 87.9\%, 60.52\%, 66.57\% on environments in higher dimensions.

\noindent{\bf Collision checking.} In Figure~\ref{figure:overall} (b), we see significant reduction of collision checking using the proposed approach, in comparison to other approaches especially in high-dimensional problems. 
The average number of collision checks by the GNN explorer is 336.3, 715.7, 2474.0, 1602.2, 350.5, 521.7, 487.0 on the environments, whereas 
BIT* needs 112\%, 175\%, 164\%, 101\%, 557\%, 226\%, 263\% times as many collision checks as our method requires on each environments. LazySP needs 105\%, 114\%, 111\%, 100\%, 105\%, 105\%, 124\% times as many collision checks as GNN requires on each environments. NEXT requires 270.23 checks on Easy2D and increases to 1206.1 on Hard2D. On the 14D environment, our method uses 17.4\% of the collision checks as what NEXT requires. 

\noindent{\bf Path cost.} We show the average path cost over all problems where all algorithms successfully found complete paths. With the smoother and obstacle encoding, our GNN approach provides the best results from UR5 to 14D, and generates comparable results for the 2D Easy and 2D Hard, where NEXT is 1.02, 1.71, and GNN is 1.18, 2.05. We find that although the GNN explorer does not yield shorter path with obstacle encoding, these explored paths can be improved further with the smoother. The reason may be that with additional obstacle encoding, the GNN explorer tends to explore edges with less probability of collision and provides more space for the smoother to improve the paths. 

\noindent{\bf Planning time.} A common concern about learning-based methods is that their running cost due to the frequent calling of a large neural network model at inference time (as seen for the NEXT curve). We see in Figure\ref{figure:overall}(d) that the wall clock time of using the GNN models is comparable to the standard heuristic-based LazySP, BIT* and RRT*, when all algorithms can find paths. The main reason is that the reduction in collision checking significantly reduces the overall time. We believe the GNNs can be further optimized to achieve faster inference as well. 

In summary, experimental results show that the GNN-based approaches  significantly reduce collision checking while maintaining high success rate, low path cost, and fast overall planning. The performance scales well from low-dimensional to high-dimensional problems.

\subsection{Generalizability of Collision Reduction}

A major challenge of learning-based approaches for planning is that small changes of the geometry of the environment can lead to abrupt change in the solutions, and thus lead the difficulty of generalization. In Figure~\ref{generalization}, we provide evidence that the GNN approach can alleviate this problem because of its access to the graph structures formed by samples uniformly taken from the space. In the 2D maze environment, the top one is easy while the bottom one is much harder, although the difference is just whether a narrow corridor is present to the left of the start state. For the UR5 with pads and poles, to solve the hard one, the robot arm needs to first rotate itself around the z-axis to bypass the small pads, and then rotate it back to fit the goal configuration. These problems are especially challenging for generalizing learned results to unseen environments. 

We observe that the GNN-based components can handle the transition from the easy to hard problems consistently. In both environments, the GNN components find paths quickly, with 9 and 236 edges explored for 2D maze, and with 1 and 256 edges explored for UR5. In contrast, the NEXT model trained on the same training sets, can quickly find a path in the easy problem, but gets stuck in the hard one. Since the two problems are close to each other in the input space for NEXT, it is not surprising to see this difficulty of generalization. In fact, the only case where NEXT can eventually find a path on 2D Maze after more than 6K edge exploration is when the algorithm delegates 10\% operations to standard RRT*. Without delegating to RRT*, NEXT gets stuck in local regions after 10K exploration steps on 2D and 1K steps on UR5. 

\begin{figure*}[ht]  
\centering
\centerline{\includegraphics[scale=0.44]{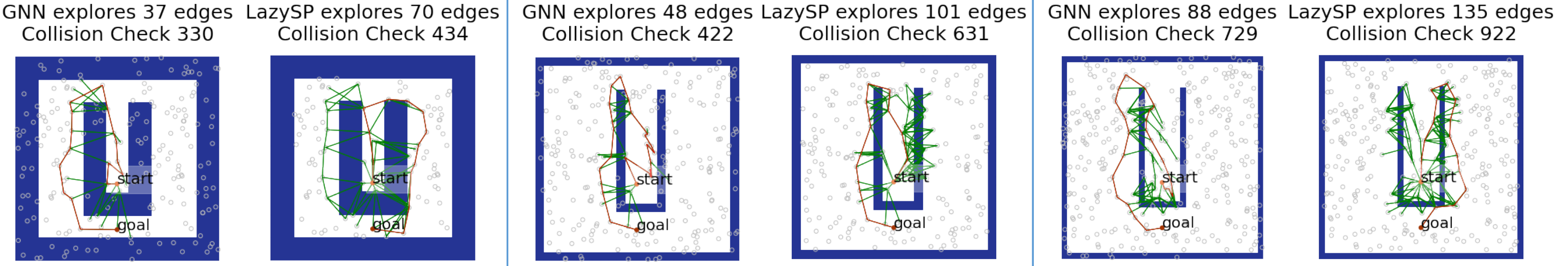}}
\caption{Comparison of performances with LazySP across different batch samples. The final path is colored red, and the other explored edges are colored green. From left to rightm we show examples of the search results with the batch size being 100, 200, and 300, respectively. We show obstacles of different thickness because the in-collision samples are also part of the inputs of the GNN planner, which can provide useful information about the topology of the environment.} 
\label{figure:lazy}
\end{figure*}

\subsection{Further Comparison with Lazy Motion Planning}

We compare GNN with the lazy motion planning method LazySP~\cite{LazySP}. Lazy approaches prioritize the collision checking on edges that are part of the shortest paths in the RGG, which is a strategy that can often see good performance especially on randomly generated graphs. However, as lazy planning uses the fixed heuristic of prioritizing certain paths, it is easy to come up with environments where this heuristic becomes misleading. Our proposed learning-based approach, instead, uses GNN to discover patterns from the training set, and can thus be viewed as a data-dependent way of forming heuristics for reducing collision checking.

Consider U-shaped obstacles, with the start state close to the bottom of the U-shape, as shown in Figure~\ref{figure:lazy}. It is a standard environment that is particularly hard for the standard lazy approach, which prioritize the edges that directly connect the start and goal states, as they are close in the RGG that does not consider obstacles. Indeed, the lazy approach needs to check most of the edges that cross the obstacles before the path for getting out of the U-shape can be found. We train the GNN-based model on these environments and observe clear benefits of learning-based components in avoiding this issue. Figure~\ref{figure:lazy} shows the difference between the two approaches in several examples of the 2D environment using different sizes of sample batches, where the GNN-based approach can typically save 50\% of collision checking on edges compared to the lazy approach. 

\subsection{Ablation Study: Probing Samples}

\begin{figure*}[ht]  
\centering
\centerline{\includegraphics[scale=0.66]{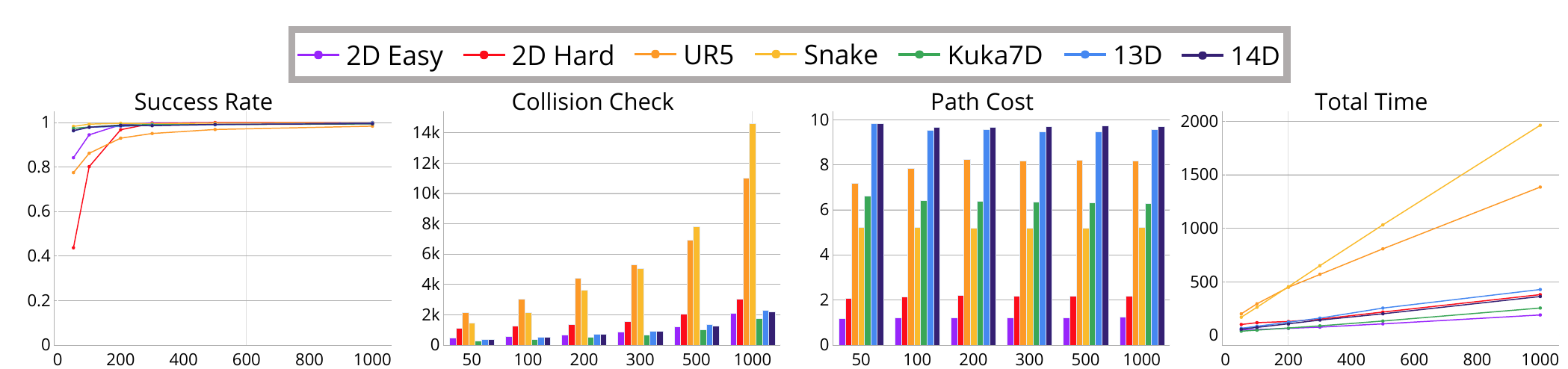}}
\caption{Comparison of performances for different probing samples from 2D to 14D problems.} 
\label{figure:probing}
\end{figure*}

We perform ablation studies of varying different parameters in our approach. {\em The full details are provided in the Appendix.} For instance, we use RGGs with different number of vertices from the free space, using 100, 200, 300, 500, 1000 samples, as illustrated in Figure \ref{figure:probing}. The success rate increases when there are more probing samples, which is consistent with the resolution-complete property of sampling-based planning. The path cost stays nearly the same for all settings, indicating robustness of the smoother models. The collision checks and planning time grows linearly with the probing samples. The reason is that the number of edges of k-NN RGG increments linearly with vertices, the input to the GNN grows linearly, thus the computation cost increases linearly on both CPU and GPU. 

\section{Conclusion}

We presented a new learning-based approach to reducing collision checking in sampling-based motion planning. We train graph neural network (GNN) models to perform path exploration and path smoothing given the random geometric graphs (RGGs) generated from batch sampling. 
We rely on the ability of GNN for capturing important geometric patterns in graphs. The learned components can significantly reduce collision checking and improve overall planning efficiency in complex and high-dimensional motion planning environments. 

\begin{ack}
This material is based upon work supported by the United States Air Force and DARPA under Contract No. FA8750-18-C-0092, AFOSR YIP FA9550-19-1-0041, NSF
Career CCF 2047034, and NSF NRI 1830399. We thank the anonymous reviewers for various important suggestions, such as for the comparison with lazy planning approaches. 
\end{ack}

\medskip

\bibliography{main}

\begin{thebibliography}{67}
\providecommand{\natexlab}[1]{#1}
\providecommand{\url}[1]{\texttt{#1}}
\expandafter\ifx\csname urlstyle\endcsname\relax
  \providecommand{\doi}[1]{doi: #1}\else
  \providecommand{\doi}{doi: \begingroup \urlstyle{rm}\Url}\fi

\bibitem[Barraquand and Latombe(1993)]{nonholonomic1}
J.~Barraquand and J.~Latombe.
\newblock Nonholonomic multibody mobile robots: Controllability and motion
  planning in the presence of obstacles.
\newblock \emph{Algorithmica}, 10\penalty0 (2-4):\penalty0 121--155, 1993.
\newblock \doi{10.1007/BF01891837}.
\newblock URL \url{https://doi.org/10.1007/BF01891837}.

\bibitem[Bency et~al.(2019)Bency, Qureshi, and Yip]{OracleNet}
M.~J. Bency, A.~H. Qureshi, and M.~C. Yip.
\newblock Neural path planning: Fixed time, near-optimal path generation via
  oracle imitation.
\newblock In \emph{2019 {IEEE/RSJ} International Conference on Intelligent
  Robots and Systems, {IROS} 2019, Macau, SAR, China, November 3-8, 2019},
  pages 3965--3972. {IEEE}, 2019.
\newblock \doi{10.1109/IROS40897.2019.8968089}.
\newblock URL \url{https://doi.org/10.1109/IROS40897.2019.8968089}.

\bibitem[Bhardwaj et~al.(2019)Bhardwaj, Choudhury, Boots, and
  Srinivasa]{lazy_experience}
M.~Bhardwaj, S.~Choudhury, B.~Boots, and S.~S. Srinivasa.
\newblock Leveraging experience in lazy search.
\newblock In A.~Bicchi, H.~Kress{-}Gazit, and S.~Hutchinson, editors,
  \emph{Robotics: Science and Systems XV, University of Freiburg, Freiburg im
  Breisgau, Germany, June 22-26, 2019}, 2019.
\newblock \doi{10.15607/RSS.2019.XV.050}.
\newblock URL \url{https://doi.org/10.15607/RSS.2019.XV.050}.

\bibitem[Bohlin and Kavraki(2000)]{LazyPRM}
R.~Bohlin and L.~E. Kavraki.
\newblock Path planning using lazy {PRM}.
\newblock In \emph{Proceedings of the 2000 {IEEE} International Conference on
  Robotics and Automation, {ICRA} 2000, April 24-28, 2000, San Francisco, CA,
  {USA}}, pages 521--528. {IEEE}, 2000.
\newblock \doi{10.1109/ROBOT.2000.844107}.
\newblock URL \url{https://doi.org/10.1109/ROBOT.2000.844107}.

\bibitem[Bresson and Laurent(2021)]{TransformerTSP}
X.~Bresson and T.~Laurent.
\newblock The transformer network for the traveling salesman problem.
\newblock \emph{CoRR}, abs/2103.03012, 2021.
\newblock URL \url{https://arxiv.org/abs/2103.03012}.

\bibitem[Chase~Kew et~al.(2021)Chase~Kew, Ichter, Bandari, Lee, and
  Faust]{Clearance}
J.~Chase~Kew, B.~Ichter, M.~Bandari, T.-W.~E. Lee, and A.~Faust.
\newblock Neural collision clearance estimator for batched motion planning.
\newblock In S.~M. LaValle, M.~Lin, T.~Ojala, D.~Shell, and J.~Yu, editors,
  \emph{Algorithmic Foundations of Robotics XIV}, pages 73--89, Cham, 2021.
  Springer International Publishing.
\newblock ISBN 978-3-030-66723-8.

\bibitem[Chazelle(1984)]{collisionNP}
B.~Chazelle.
\newblock Convex partitions of polyhedra: {A} lower bound and worst-case
  optimal algorithm.
\newblock \emph{{SIAM} J. Comput.}, 13\penalty0 (3):\penalty0 488--507, 1984.
\newblock \doi{10.1137/0213031}.
\newblock URL \url{https://doi.org/10.1137/0213031}.

\bibitem[Chen et~al.(2020)Chen, Dai, Lin, Ye, Liu, and Song]{NEXT}
B.~Chen, B.~Dai, Q.~Lin, G.~Ye, H.~Liu, and L.~Song.
\newblock Learning to plan in high dimensions via neural
  exploration-exploitation trees.
\newblock In \emph{8th International Conference on Learning Representations,
  {ICLR} 2020, Addis Ababa, Ethiopia, April 26-30, 2020}. OpenReview.net, 2020.
\newblock URL \url{https://openreview.net/forum?id=rJgJDAVKvB}.

\bibitem[Cherif(1999)]{kino1}
M.~Cherif.
\newblock Kinodynamic motion planning for all-terrain wheeled vehicles.
\newblock In \emph{1999 {IEEE} International Conference on Robotics and
  Automation, Marriott Hotel, Renaissance Center, Detroit, Michigan, USA, May
  10-15, 1999, Proceedings}, pages 317--322. {IEEE} Robotics and Automation
  Society, 1999.
\newblock \doi{10.1109/ROBOT.1999.769998}.
\newblock URL \url{https://doi.org/10.1109/ROBOT.1999.769998}.

\bibitem[Choset et~al.(2005)Choset, Lynch, Hutchinson, Kantor, Burgard,
  Kavraki, Thrun, and Arkin]{choset2005principles}
H.~M. Choset, K.~M. Lynch, S.~Hutchinson, G.~Kantor, W.~Burgard, L.~Kavraki,
  S.~Thrun, and R.~C. Arkin.
\newblock \emph{Principles of robot motion: theory, algorithms, and
  implementation}.
\newblock MIT press, 2005.

\bibitem[Cohen et~al.(2014)Cohen, Phillips, and Likhachev]{LWA}
B.~J. Cohen, M.~Phillips, and M.~Likhachev.
\newblock Planning single-arm manipulations with n-arm robots.
\newblock In D.~Fox, L.~E. Kavraki, and H.~Kurniawati, editors, \emph{Robotics:
  Science and Systems X, University of California, Berkeley, USA, July 12-16,
  2014}, 2014.
\newblock \doi{10.15607/RSS.2014.X.033}.
\newblock URL \url{http://www.roboticsproceedings.org/rss10/p33.html}.

\bibitem[Dai et~al.(2019)Dai, Li, Wang, Singh, Huang, and Kohli]{GraphExplore}
H.~Dai, Y.~Li, C.~Wang, R.~Singh, P.~Huang, and P.~Kohli.
\newblock Learning transferable graph exploration.
\newblock In H.~M. Wallach, H.~Larochelle, A.~Beygelzimer,
  F.~d'Alch{\'{e}}{-}Buc, E.~B. Fox, and R.~Garnett, editors, \emph{Advances in
  Neural Information Processing Systems 32: Annual Conference on Neural
  Information Processing Systems 2019, NeurIPS 2019, December 8-14, 2019,
  Vancouver, BC, Canada}, pages 2514--2525, 2019.
\newblock URL
  \url{https://proceedings.neurips.cc/paper/2019/hash/afe434653a898da20044041262b3ac74-Abstract.html}.

\bibitem[Das and Yip(2020)]{Fastron}
N.~Das and M.~Yip.
\newblock Learning-based proxy collision detection for robot motion planning
  applications.
\newblock \emph{IEEE Transactions on Robotics}, 36\penalty0 (4):\penalty0
  1096--1114, 2020.

\bibitem[Diankov and Kuffner(2007)]{randomA}
R.~Diankov and J.~Kuffner.
\newblock Randomized statistical path planning.
\newblock In \emph{2007 {IEEE/RSJ} International Conference on Intelligent
  Robots and Systems, October 29 - November 2, 2007, Sheraton Hotel and Marina,
  San Diego, California, {USA}}, pages 1--6. {IEEE}, 2007.
\newblock \doi{10.1109/IROS.2007.4399557}.
\newblock URL \url{https://doi.org/10.1109/IROS.2007.4399557}.

\bibitem[Donald et~al.(1993)Donald, Xavier, Canny, and Reif]{kino2}
B.~R. Donald, P.~G. Xavier, J.~F. Canny, and J.~H. Reif.
\newblock Kinodynamic motion planning.
\newblock \emph{J. {ACM}}, 40\penalty0 (5):\penalty0 1048--1066, 1993.
\newblock \doi{10.1145/174147.174150}.
\newblock URL \url{https://doi.org/10.1145/174147.174150}.

\bibitem[Drori et~al.(2020)Drori, Kharkar, Sickinger, Kates, Ma, Ge, Dolev,
  Dietrich, Williamson, and Udell]{CombGraphLinear}
I.~Drori, A.~Kharkar, W.~R. Sickinger, B.~Kates, Q.~Ma, S.~Ge, E.~Dolev,
  B.~Dietrich, D.~P. Williamson, and M.~Udell.
\newblock Learning to solve combinatorial optimization problems on real-world
  graphs in linear time.
\newblock In M.~A. Wani, F.~Luo, X.~A. Li, D.~Dou, and F.~Bonchi, editors,
  \emph{19th {IEEE} International Conference on Machine Learning and
  Applications, {ICMLA} 2020, Miami, FL, USA, December 14-17, 2020}, pages
  19--24. {IEEE}, 2020.
\newblock \doi{10.1109/ICMLA51294.2020.00013}.
\newblock URL \url{https://doi.org/10.1109/ICMLA51294.2020.00013}.

\bibitem[Emmons et~al.(2020)Emmons, Jain, Laskin, Kurutach, Abbeel, and
  Pathak]{SGM}
S.~Emmons, A.~Jain, M.~Laskin, T.~Kurutach, P.~Abbeel, and D.~Pathak.
\newblock Sparse graphical memory for robust planning.
\newblock In H.~Larochelle, M.~Ranzato, R.~Hadsell, M.~Balcan, and H.~Lin,
  editors, \emph{Advances in Neural Information Processing Systems 33: Annual
  Conference on Neural Information Processing Systems 2020, NeurIPS 2020,
  December 6-12, 2020, virtual}, 2020.
\newblock URL
  \url{https://proceedings.neurips.cc/paper/2020/hash/385822e359afa26d52b5b286226f2cea-Abstract.html}.

\bibitem[Eysenbach et~al.(2019)Eysenbach, Salakhutdinov, and Levine]{SoRB}
B.~Eysenbach, R.~Salakhutdinov, and S.~Levine.
\newblock Search on the replay buffer: Bridging planning and reinforcement
  learning.
\newblock In H.~M. Wallach, H.~Larochelle, A.~Beygelzimer,
  F.~d'Alch{\'{e}}{-}Buc, E.~B. Fox, and R.~Garnett, editors, \emph{Advances in
  Neural Information Processing Systems 32: Annual Conference on Neural
  Information Processing Systems 2019, NeurIPS 2019, December 8-14, 2019,
  Vancouver, BC, Canada}, pages 15220--15231, 2019.
\newblock URL
  \url{https://proceedings.neurips.cc/paper/2019/hash/5c48ff18e0a47baaf81d8b8ea51eec92-Abstract.html}.

\bibitem[Gammell et~al.(2014)Gammell, Srinivasa, and Barfoot]{bit}
J.~D. Gammell, S.~S. Srinivasa, and T.~D. Barfoot.
\newblock Bit*: Batch informed trees for optimal sampling-based planning via
  dynamic programming on implicit random geometric graphs.
\newblock \emph{CoRR}, abs/1405.5848, 2014.
\newblock URL \url{http://arxiv.org/abs/1405.5848}.

\bibitem[Gilbert(1961)]{rdisc}
E.~N. Gilbert.
\newblock Random plane networks.
\newblock \emph{Journal of the society for industrial and applied mathematics},
  9\penalty0 (4):\penalty0 533--543, 1961.

\bibitem[Haghtalab et~al.(2019)Haghtalab, Mackenzie, Procaccia, Salzman, and
  Srinivasa]{LazySP}
N.~Haghtalab, S.~Mackenzie, A.~D. Procaccia, O.~Salzman, and S.~S. Srinivasa.
\newblock The provable virtue of laziness in motion planning.
\newblock In S.~Kraus, editor, \emph{Proceedings of the Twenty-Eighth
  International Joint Conference on Artificial Intelligence, {IJCAI} 2019,
  Macao, China, August 10-16, 2019}, pages 6161--6165. ijcai.org, 2019.
\newblock \doi{10.24963/ijcai.2019/855}.
\newblock URL \url{https://doi.org/10.24963/ijcai.2019/855}.

\bibitem[Heiden et~al.(2018)Heiden, Palmieri, Koenig, Arras, and
  Sukhatme]{GradSmooth}
E.~Heiden, L.~Palmieri, S.~Koenig, K.~O. Arras, and G.~S. Sukhatme.
\newblock Gradient-informed path smoothing for wheeled mobile robots.
\newblock In \emph{2018 {IEEE} International Conference on Robotics and
  Automation, {ICRA} 2018, Brisbane, Australia, May 21-25, 2018}, pages
  1710--1717. {IEEE}, 2018.
\newblock \doi{10.1109/ICRA.2018.8460818}.
\newblock URL \url{https://doi.org/10.1109/ICRA.2018.8460818}.

\bibitem[Hochreiter and Schmidhuber(1997)]{LSTM}
S.~Hochreiter and J.~Schmidhuber.
\newblock {Long Short-Term Memory}.
\newblock \emph{Neural Computation}, 9\penalty0 (8):\penalty0 1735--1780, 11
  1997.
\newblock ISSN 0899-7667.
\newblock \doi{10.1162/neco.1997.9.8.1735}.
\newblock URL \url{https://doi.org/10.1162/neco.1997.9.8.1735}.

\bibitem[Ichter and Pavone(2019)]{L2RRT}
B.~Ichter and M.~Pavone.
\newblock Robot motion planning in learned latent spaces.
\newblock \emph{{IEEE} Robotics Autom. Lett.}, 4\penalty0 (3):\penalty0
  2407--2414, 2019.
\newblock \doi{10.1109/LRA.2019.2901898}.
\newblock URL \url{https://doi.org/10.1109/LRA.2019.2901898}.

\bibitem[Ichter et~al.(2018)Ichter, Harrison, and Pavone]{CVAE}
B.~Ichter, J.~Harrison, and M.~Pavone.
\newblock Learning sampling distributions for robot motion planning.
\newblock In \emph{2018 {IEEE} International Conference on Robotics and
  Automation, {ICRA} 2018, Brisbane, Australia, May 21-25, 2018}, pages
  7087--7094. {IEEE}, 2018.
\newblock \doi{10.1109/ICRA.2018.8460730}.
\newblock URL \url{https://doi.org/10.1109/ICRA.2018.8460730}.

\bibitem[Ioffe and Szegedy(2015)]{BatchNorm}
S.~Ioffe and C.~Szegedy.
\newblock Batch normalization: Accelerating deep network training by reducing
  internal covariate shift.
\newblock In \emph{International conference on machine learning}, pages
  448--456. PMLR, 2015.

\bibitem[Janson and Pavone(2013)]{DBLP:journals/corr/JansonP13}
L.~Janson and M.~Pavone.
\newblock Fast marching trees: a fast marching sampling-based method for
  optimal motion planning in many dimensions - extended version.
\newblock \emph{CoRR}, abs/1306.3532, 2013.
\newblock URL \url{http://arxiv.org/abs/1306.3532}.

\bibitem[Jim{\'e}nez et~al.(1998)Jim{\'e}nez, Thomas, and
  Torras]{CollisionBook}
P.~Jim{\'e}nez, F.~Thomas, and C.~Torras.
\newblock Collision detection algorithms for motion planning.
\newblock In \emph{Robot motion planning and control}, pages 305--343.
  Springer, 1998.

\bibitem[Jurgenson and Tamar(2019)]{RSS}
T.~Jurgenson and A.~Tamar.
\newblock Harnessing reinforcement learning for neural motion planning.
\newblock In A.~Bicchi, H.~Kress{-}Gazit, and S.~Hutchinson, editors,
  \emph{Robotics: Science and Systems XV, University of Freiburg, Freiburg im
  Breisgau, Germany, June 22-26, 2019}, 2019.
\newblock \doi{10.15607/RSS.2019.XV.026}.
\newblock URL \url{https://doi.org/10.15607/RSS.2019.XV.026}.

\bibitem[Kalakrishnan et~al.(2011)Kalakrishnan, Chitta, Theodorou, Pastor, and
  Schaal]{stomp}
M.~Kalakrishnan, S.~Chitta, E.~Theodorou, P.~Pastor, and S.~Schaal.
\newblock Stomp: Stochastic trajectory optimization for motion planning.
\newblock In \emph{2011 IEEE international conference on robotics and
  automation}, pages 4569--4574. IEEE, 2011.

\bibitem[Karaman and Frazzoli(2011)]{RRT*}
S.~Karaman and E.~Frazzoli.
\newblock Sampling-based algorithms for optimal motion planning.
\newblock \emph{Int. J. Robotics Res.}, 30\penalty0 (7):\penalty0 846--894,
  2011.
\newblock \doi{10.1177/0278364911406761}.
\newblock URL \url{https://doi.org/10.1177/0278364911406761}.

\bibitem[Khalil et~al.(2017)Khalil, Dai, Zhang, Dilkina, and Song]{CombGraph}
E.~B. Khalil, H.~Dai, Y.~Zhang, B.~Dilkina, and L.~Song.
\newblock Learning combinatorial optimization algorithms over graphs.
\newblock In I.~Guyon, U.~von Luxburg, S.~Bengio, H.~M. Wallach, R.~Fergus,
  S.~V.~N. Vishwanathan, and R.~Garnett, editors, \emph{Advances in Neural
  Information Processing Systems 30: Annual Conference on Neural Information
  Processing Systems 2017, December 4-9, 2017, Long Beach, CA, {USA}}, pages
  6348--6358, 2017.
\newblock URL
  \url{https://proceedings.neurips.cc/paper/2017/hash/d9896106ca98d3d05b8cbdf4fd8b13a1-Abstract.html}.

\bibitem[Khan et~al.(2020)Khan, Ribeiro, Kumar, and Francis]{GNNMP}
A.~Khan, A.~Ribeiro, V.~Kumar, and A.~G. Francis.
\newblock Graph neural networks for motion planning.
\newblock \emph{CoRR}, abs/2006.06248, 2020.
\newblock URL \url{https://arxiv.org/abs/2006.06248}.

\bibitem[Kondo(1991)]{mani1}
K.~Kondo.
\newblock Motion planning with six degrees of freedom by multistrategic
  bidirectional heuristic free-space enumeration.
\newblock \emph{{IEEE} Trans. Robotics Autom.}, 7\penalty0 (3):\penalty0
  267--277, 1991.
\newblock \doi{10.1109/70.88136}.
\newblock URL \url{https://doi.org/10.1109/70.88136}.

\bibitem[LaValle(2006)]{DBLP:books/cu/L2006}
S.~M. LaValle.
\newblock \emph{Planning Algorithms}.
\newblock Cambridge University Press, 2006.
\newblock ISBN 9780511546877.
\newblock \doi{10.1017/CBO9780511546877}.
\newblock URL \url{http://planning.cs.uiuc.edu/}.

\bibitem[LaValle and Jr.(1999)]{rrt}
S.~M. LaValle and J.~J.~K. Jr.
\newblock Randomized kinodynamic planning.
\newblock In \emph{1999 {IEEE} International Conference on Robotics and
  Automation, Marriott Hotel, Renaissance Center, Detroit, Michigan, USA, May
  10-15, 1999, Proceedings}, pages 473--479. {IEEE} Robotics and Automation
  Society, 1999.
\newblock \doi{10.1109/ROBOT.1999.770022}.
\newblock URL \url{https://doi.org/10.1109/ROBOT.1999.770022}.

\bibitem[Lee et~al.(2018)Lee, Parisotto, Chaplot, Xing, and
  Salakhutdinov]{GPPN}
L.~Lee, E.~Parisotto, D.~S. Chaplot, E.~P. Xing, and R.~Salakhutdinov.
\newblock Gated path planning networks.
\newblock In J.~G. Dy and A.~Krause, editors, \emph{Proceedings of the 35th
  International Conference on Machine Learning, {ICML} 2018,
  Stockholmsm{\"{a}}ssan, Stockholm, Sweden, July 10-15, 2018}, volume~80 of
  \emph{Proceedings of Machine Learning Research}, pages 2953--2961. {PMLR},
  2018.
\newblock URL \url{http://proceedings.mlr.press/v80/lee18c.html}.

\bibitem[Li et~al.(2019)Li, Gama, Ribeiro, and Prorok]{decentral}
Q.~Li, F.~Gama, A.~Ribeiro, and A.~Prorok.
\newblock Graph neural networks for decentralized multi-robot path planning.
\newblock \emph{CoRR}, abs/1912.06095, 2019.
\newblock URL \url{http://arxiv.org/abs/1912.06095}.

\bibitem[Lynch and Mason(1996)]{nonholonomic2}
K.~M. Lynch and M.~T. Mason.
\newblock Stable pushing: Mechanics, controllability, and planning.
\newblock \emph{Int. J. Robotics Res.}, 15\penalty0 (6):\penalty0 533--556,
  1996.
\newblock \doi{10.1177/027836499601500602}.
\newblock URL \url{https://doi.org/10.1177/027836499601500602}.

\bibitem[Mandalika et~al.(2018)Mandalika, Salzman, and Srinivasa]{LRA}
A.~Mandalika, O.~Salzman, and S.~S. Srinivasa.
\newblock Lazy receding horizon a* for efficient path planning in graphs with
  expensive-to-evaluate edges.
\newblock In M.~de~Weerdt, S.~Koenig, G.~R{\"{o}}ger, and M.~T.~J. Spaan,
  editors, \emph{Proceedings of the Twenty-Eighth International Conference on
  Automated Planning and Scheduling, {ICAPS} 2018, Delft, The Netherlands, June
  24-29, 2018}, pages 476--484. {AAAI} Press, 2018.
\newblock URL
  \url{https://aaai.org/ocs/index.php/ICAPS/ICAPS18/paper/view/17785}.

\bibitem[Mandalika et~al.(2019)Mandalika, Choudhury, Salzman, and
  Srinivasa]{gls}
A.~Mandalika, S.~Choudhury, O.~Salzman, and S.~S. Srinivasa.
\newblock Generalized lazy search for robot motion planning: Interleaving
  search and edge evaluation via event-based toggles.
\newblock In J.~Benton, N.~Lipovetzky, E.~Onaindia, D.~E. Smith, and
  S.~Srivastava, editors, \emph{Proceedings of the Twenty-Ninth International
  Conference on Automated Planning and Scheduling, {ICAPS} 2018, Berkeley, CA,
  USA, July 11-15, 2019}, pages 745--753. {AAAI} Press, 2019.
\newblock URL \url{https://aaai.org/ojs/index.php/ICAPS/article/view/3543}.

\bibitem[Mueller et~al.(2015)Mueller, Hehn, and
  D'Andrea]{mueller2015computationally}
M.~W. Mueller, M.~Hehn, and R.~D'Andrea.
\newblock A computationally efficient motion primitive for quadrocopter
  trajectory generation.
\newblock \emph{IEEE Transactions on Robotics}, 31\penalty0 (6):\penalty0
  1294--1310, 2015.

\bibitem[Niu et~al.(2018)Niu, Chen, Guo, Targonski, Smith, and Kovacevic]{GVIN}
S.~Niu, S.~Chen, H.~Guo, C.~Targonski, M.~C. Smith, and J.~Kovacevic.
\newblock Generalized value iteration networks: Life beyond lattices.
\newblock In S.~A. McIlraith and K.~Q. Weinberger, editors, \emph{Proceedings
  of the Thirty-Second {AAAI} Conference on Artificial Intelligence, (AAAI-18),
  the 30th innovative Applications of Artificial Intelligence (IAAI-18), and
  the 8th {AAAI} Symposium on Educational Advances in Artificial Intelligence
  (EAAI-18), New Orleans, Louisiana, USA, February 2-7, 2018}, pages
  6246--6253. {AAAI} Press, 2018.
\newblock URL
  \url{https://www.aaai.org/ocs/index.php/AAAI/AAAI18/paper/view/16552}.

\bibitem[Persson and Sharf(2014)]{DBLP:journals/ijrr/PerssonS14}
S.~M. Persson and I.~Sharf.
\newblock Sampling-based a* algorithm for robot path-planning.
\newblock \emph{Int. J. Robotics Res.}, 33\penalty0 (13):\penalty0 1683--1708,
  2014.
\newblock \doi{10.1177/0278364914547786}.
\newblock URL \url{https://doi.org/10.1177/0278364914547786}.

\bibitem[Qi et~al.(2017)Qi, Su, Mo, and Guibas]{PointNet}
C.~R. Qi, H.~Su, K.~Mo, and L.~J. Guibas.
\newblock Pointnet: Deep learning on point sets for 3d classification and
  segmentation.
\newblock In \emph{2017 {IEEE} Conference on Computer Vision and Pattern
  Recognition, {CVPR} 2017, Honolulu, HI, USA, July 21-26, 2017}, pages 77--85.
  {IEEE} Computer Society, 2017.
\newblock \doi{10.1109/CVPR.2017.16}.
\newblock URL \url{https://doi.org/10.1109/CVPR.2017.16}.

\bibitem[Qureshi et~al.(2021)Qureshi, Miao, Simeonov, and Yip]{MPNet}
A.~H. Qureshi, Y.~Miao, A.~Simeonov, and M.~C. Yip.
\newblock Motion planning networks: Bridging the gap between learning-based and
  classical motion planners.
\newblock \emph{IEEE Transactions on Robotics}, 37\penalty0 (1):\penalty0
  48--66, 2021.
\newblock \doi{10.1109/TRO.2020.3006716}.

\bibitem[{Reif}(1979)]{pspace-hard}
J.~H. {Reif}.
\newblock Complexity of the mover's problem and generalizations.
\newblock In \emph{20th Annual Symposium on Foundations of Computer Science
  (sfcs 1979)}, pages 421--427, 1979.
\newblock \doi{10.1109/SFCS.1979.10}.

\bibitem[Rendle et~al.(2009)Rendle, Freudenthaler, Gantner, and
  Schmidt{-}Thieme]{BPR}
S.~Rendle, C.~Freudenthaler, Z.~Gantner, and L.~Schmidt{-}Thieme.
\newblock {BPR:} bayesian personalized ranking from implicit feedback.
\newblock In J.~A. Bilmes and A.~Y. Ng, editors, \emph{{UAI} 2009, Proceedings
  of the Twenty-Fifth Conference on Uncertainty in Artificial Intelligence,
  Montreal, QC, Canada, June 18-21, 2009}, pages 452--461. {AUAI} Press, 2009.
\newblock URL
  \url{https://dslpitt.org/uai/displayArticleDetails.jsp?mmnu=1\&smnu=2\&article\_id=1630\&proceeding\_id=25}.

\bibitem[Savinov et~al.(2018)Savinov, Dosovitskiy, and Koltun]{SPTM}
N.~Savinov, A.~Dosovitskiy, and V.~Koltun.
\newblock Semi-parametric topological memory for navigation.
\newblock In \emph{6th International Conference on Learning Representations,
  {ICLR} 2018, Vancouver, BC, Canada, April 30 - May 3, 2018, Conference Track
  Proceedings}. OpenReview.net, 2018.
\newblock URL \url{https://openreview.net/forum?id=SygwwGbRW}.

\bibitem[Schulman et~al.(2014)Schulman, Duan, Ho, Lee, Awwal, Bradlow, Pan,
  Patil, Goldberg, and Abbeel]{trajopt}
J.~Schulman, Y.~Duan, J.~Ho, A.~Lee, I.~Awwal, H.~Bradlow, J.~Pan, S.~Patil,
  K.~Goldberg, and P.~Abbeel.
\newblock Motion planning with sequential convex optimization and convex
  collision checking.
\newblock \emph{The International Journal of Robotics Research}, 33\penalty0
  (9):\penalty0 1251--1270, 2014.

\bibitem[Shah et~al.(2021)Shah, Eysenbach, Rhinehart, and Levine]{RECON}
D.~Shah, B.~Eysenbach, N.~Rhinehart, and S.~Levine.
\newblock {RECON:} rapid exploration for open-world navigation with latent goal
  models.
\newblock \emph{CoRR}, abs/2104.05859, 2021.
\newblock URL \url{https://arxiv.org/abs/2104.05859}.

\bibitem[Strub and Gammell(2020)]{ait}
M.~P. Strub and J.~D. Gammell.
\newblock Advanced bit* (abit*): Sampling-based planning with advanced
  graph-search techniques.
\newblock In \emph{2020 {IEEE} International Conference on Robotics and
  Automation, {ICRA} 2020, Paris, France, May 31 - August 31, 2020}, pages
  130--136. {IEEE}, 2020.
\newblock \doi{10.1109/ICRA40945.2020.9196580}.
\newblock URL \url{https://doi.org/10.1109/ICRA40945.2020.9196580}.

\bibitem[Strudel et~al.(2020)Strudel, Garcia, Carpentier, Laumond, Laptev, and
  Schmid]{Obstacle}
R.~A.~M. Strudel, R.~Garcia, J.~Carpentier, J.~Laumond, I.~Laptev, and
  C.~Schmid.
\newblock Learning obstacle representations for neural motion planning.
\newblock \emph{CoRR}, abs/2008.11174, 2020.
\newblock URL \url{https://arxiv.org/abs/2008.11174}.

\bibitem[Tamar et~al.(2016)Tamar, Levine, Abbeel, Wu, and Thomas]{VIN}
A.~Tamar, S.~Levine, P.~Abbeel, Y.~Wu, and G.~Thomas.
\newblock Value iteration networks.
\newblock In D.~D. Lee, M.~Sugiyama, U.~von Luxburg, I.~Guyon, and R.~Garnett,
  editors, \emph{Advances in Neural Information Processing Systems 29: Annual
  Conference on Neural Information Processing Systems 2016, December 5-10,
  2016, Barcelona, Spain}, pages 2146--2154, 2016.
\newblock URL
  \url{https://proceedings.neurips.cc/paper/2016/hash/c21002f464c5fc5bee3b98ced83963b8-Abstract.html}.

\bibitem[Tang et~al.(2020)Tang, Huang, Gu, Lu, and Su]{HomoIterGNN}
H.~Tang, Z.~Huang, J.~Gu, B.-L. Lu, and H.~Su.
\newblock Towards scale-invariant graph-related problem solving by iterative
  homogeneous gnns.
\newblock In H.~Larochelle, M.~Ranzato, R.~Hadsell, M.~F. Balcan, and H.~Lin,
  editors, \emph{Advances in Neural Information Processing Systems}, volume~33,
  pages 15811--15822. Curran Associates, Inc., 2020.
\newblock URL
  \url{https://proceedings.neurips.cc/paper/2020/file/b64a70760bb75e3ecfd1ad86d8f10c88-Paper.pdf}.

\bibitem[Toussaint(2015)]{logicgeometric}
M.~Toussaint.
\newblock Logic-geometric programming: An optimization-based approach to
  combined task and motion planning.
\newblock In \emph{IJCAI}, pages 1930--1936, 2015.

\bibitem[Vaswani et~al.(2017)Vaswani, Shazeer, Parmar, Uszkoreit, Jones, Gomez,
  Kaiser, and Polosukhin]{Attention}
A.~Vaswani, N.~Shazeer, N.~Parmar, J.~Uszkoreit, L.~Jones, A.~N. Gomez,
  L.~Kaiser, and I.~Polosukhin.
\newblock Attention is all you need.
\newblock In I.~Guyon, U.~von Luxburg, S.~Bengio, H.~M. Wallach, R.~Fergus,
  S.~V.~N. Vishwanathan, and R.~Garnett, editors, \emph{Advances in Neural
  Information Processing Systems 30: Annual Conference on Neural Information
  Processing Systems 2017, December 4-9, 2017, Long Beach, CA, {USA}}, pages
  5998--6008, 2017.
\newblock URL
  \url{https://proceedings.neurips.cc/paper/2017/hash/3f5ee243547dee91fbd053c1c4a845aa-Abstract.html}.

\bibitem[Velickovic et~al.(2020{\natexlab{a}})Velickovic, Buesing, Overlan,
  Pascanu, Vinyals, and Blundell]{PointerGNN}
P.~Velickovic, L.~Buesing, M.~C. Overlan, R.~Pascanu, O.~Vinyals, and
  C.~Blundell.
\newblock Pointer graph networks.
\newblock In H.~Larochelle, M.~Ranzato, R.~Hadsell, M.~Balcan, and H.~Lin,
  editors, \emph{Advances in Neural Information Processing Systems 33: Annual
  Conference on Neural Information Processing Systems 2020, NeurIPS 2020,
  December 6-12, 2020, virtual}, 2020{\natexlab{a}}.
\newblock URL
  \url{https://proceedings.neurips.cc/paper/2020/hash/176bf6219855a6eb1f3a30903e34b6fb-Abstract.html}.

\bibitem[Velickovic et~al.(2020{\natexlab{b}})Velickovic, Ying, Padovano,
  Hadsell, and Blundell]{NeuralExe}
P.~Velickovic, R.~Ying, M.~Padovano, R.~Hadsell, and C.~Blundell.
\newblock Neural execution of graph algorithms.
\newblock In \emph{8th International Conference on Learning Representations,
  {ICLR} 2020, Addis Ababa, Ethiopia, April 26-30, 2020}. OpenReview.net,
  2020{\natexlab{b}}.
\newblock URL \url{https://openreview.net/forum?id=SkgKO0EtvS}.

\bibitem[Xu et~al.(2020)Xu, Li, Zhang, Du, Kawarabayashi, and
  Jegelka]{GNNReason}
K.~Xu, J.~Li, M.~Zhang, S.~S. Du, K.~Kawarabayashi, and S.~Jegelka.
\newblock What can neural networks reason about?
\newblock In \emph{8th International Conference on Learning Representations,
  {ICLR} 2020, Addis Ababa, Ethiopia, April 26-30, 2020}. OpenReview.net, 2020.
\newblock URL \url{https://openreview.net/forum?id=rJxbJeHFPS}.

\bibitem[Xue and Kumar(2004)]{k-NN}
F.~Xue and P.~R. Kumar.
\newblock The number of neighbors needed for connectivity of wireless networks.
\newblock \emph{Wireless networks}, 10\penalty0 (2):\penalty0 169--181, 2004.

\bibitem[Yan et~al.(2020)Yan, Swersky, Koutra, Ranganathan, and
  Hashemi]{GNNEngine}
Y.~Yan, K.~Swersky, D.~Koutra, P.~Ranganathan, and M.~Hashemi.
\newblock Neural execution engines: Learning to execute subroutines.
\newblock In H.~Larochelle, M.~Ranzato, R.~Hadsell, M.~Balcan, and H.~Lin,
  editors, \emph{Advances in Neural Information Processing Systems 33: Annual
  Conference on Neural Information Processing Systems 2020, NeurIPS 2020,
  December 6-12, 2020, virtual}, 2020.
\newblock URL
  \url{https://proceedings.neurips.cc/paper/2020/hash/c8b9abffb45bf79a630fb613dcd23449-Abstract.html}.

\bibitem[Yang et~al.(2020)Yang, Zhang, Morcos, Pineau, Abbeel, and
  Calandra]{plan2vec}
G.~Yang, A.~Zhang, A.~S. Morcos, J.~Pineau, P.~Abbeel, and R.~Calandra.
\newblock Plan2vec: Unsupervised representation learning by latent plans.
\newblock In A.~M. Bayen, A.~Jadbabaie, G.~J. Pappas, P.~A. Parrilo, B.~Recht,
  C.~J. Tomlin, and M.~N. Zeilinger, editors, \emph{Proceedings of the 2nd
  Annual Conference on Learning for Dynamics and Control, {L4DC} 2020, Online
  Event, Berkeley, CA, USA, 11-12 June 2020}, volume 120 of \emph{Proceedings
  of Machine Learning Research}, pages 935--946. {PMLR}, 2020.
\newblock URL \url{http://proceedings.mlr.press/v120/yang20b.html}.

\bibitem[Zhang et~al.(2018)Zhang, Huh, and Lee]{Implicit}
C.~Zhang, J.~Huh, and D.~D. Lee.
\newblock Learning implicit sampling distributions for motion planning.
\newblock In \emph{2018 {IEEE/RSJ} International Conference on Intelligent
  Robots and Systems, {IROS} 2018, Madrid, Spain, October 1-5, 2018}, pages
  3654--3661. {IEEE}, 2018.
\newblock \doi{10.1109/IROS.2018.8594028}.
\newblock URL \url{https://doi.org/10.1109/IROS.2018.8594028}.

\bibitem[Zhou et~al.(2019)Zhou, Gao, Wang, Liu, and Shen]{zhou2019robust}
B.~Zhou, F.~Gao, L.~Wang, C.~Liu, and S.~Shen.
\newblock Robust and efficient quadrotor trajectory generation for fast
  autonomous flight.
\newblock \emph{IEEE Robotics and Automation Letters}, 4\penalty0 (4):\penalty0
  3529--3536, 2019.

\bibitem[Zhu et~al.(2015)Zhu, Schmerling, and Pavone]{zhu2015convex}
Z.~Zhu, E.~Schmerling, and M.~Pavone.
\newblock A convex optimization approach to smooth trajectories for motion
  planning with car-like robots.
\newblock In \emph{2015 54th IEEE conference on decision and control (CDC)},
  pages 835--842. IEEE, 2015.

\bibitem[Zucker et~al.(2013)Zucker, Ratliff, Dragan, Pivtoraiko, Klingensmith,
  Dellin, Bagnell, and Srinivasa]{chomp}
M.~Zucker, N.~Ratliff, A.~D. Dragan, M.~Pivtoraiko, M.~Klingensmith, C.~M.
  Dellin, J.~A. Bagnell, and S.~S. Srinivasa.
\newblock Chomp: Covariant hamiltonian optimization for motion planning.
\newblock \emph{The International Journal of Robotics Research}, 32\penalty0
  (9-10):\penalty0 1164--1193, 2013.

\end{thebibliography}


\ifarXiv
    \foreach \x in {1,...,9}
    {
        \clearpage
        \includepdf[pages={\x}]{\supplementfilename}
    }
\fi

\end{document}











\maketitle 

\section{More Details on GNN Architectures}

\begin{figure*}[h!]  
\centering
\includegraphics[width=0.5\textwidth]{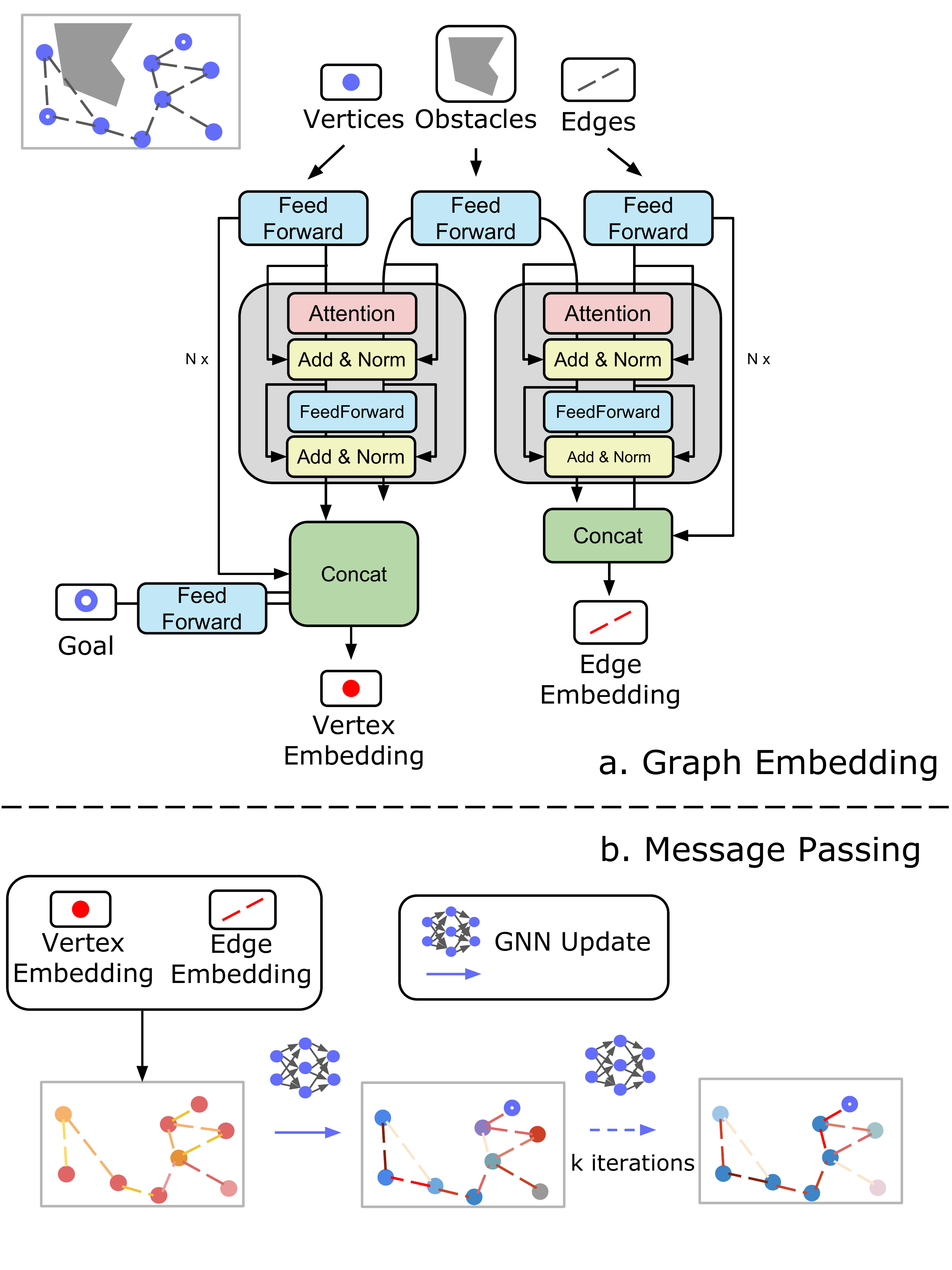}

\end{figure*}

\subsection{Obstacle Encoding}

In the experiment part, we find obstacle encoding is helpful to the GNN explorer, which can optimize the explored path further with the smoother. We elaborate on the formulation of the obstacle encoding.

In this work, we consider an obstacle as a 2D or 3D box depending on the workspace,  denoted as $o=(p_1, \cdots, p_n, l_1, \cdots, l_n)\in \mathbb{R}^{2n}, n\in [2,3]$, where $p_i$ and $l_i$ are the center and length of the box along the $i$-th dimension. The environment configuration is written as $O\in \mathbb{R}^{|\{o\}|\times 2n}$, where $|\{o\}|$ is the number of the obstacles. Note that $|\{o\}|$ is a variable number, since the number of obstacles can be different for each problem.

Given MLPs $f_{a_x}^{(i)},f_{a_y}^{(i)},f_{K_x}^{(i)},f_{Q_x}^{(i)},f_{V_x}^{(i)},f_{K_y}^{(i)},f_{Q_y}^{(i)},f_{V_y}^{(i)}$, the obstacle encoding is formulated as: 

\begin{equation}
\begin{split}
a_x&=\text{LN}(x+Att(f_{K^{(i)}_x}(O), f_{Q^{(i)}_x}(x), f_{V^{(i)}_x}(O))) \mbox{ and } x = \text{LN}(a_x + f_{a_x}^{(i)}(a_x))\\
a_y&=\text{LN}(y+Att(f_{K^{(i)}_y}(O), f_{Q^{(i)}_y}(y), f_{V^{(i)}_y}(O))) \mbox{ and }
y = \text{LN}(a_y + f_{a_y}^{(i)}(a_y))
\label{obsenc}
\end{split}
\end{equation}

where $\text{LN}$ denotes the layer normalization~\cite{LayerNorm}. This architecture follows the standard transformer block design~\cite{Transformer}.

\subsection{Special Features}

The overall special features for GNN explorer and smoother are described as follows:

\paragraph{Special features in explorer architecture.} The GNN explorer embeds vertices with $x=h_x(v, v_g, (v-v_g)^2, v-v_g)$, with an MLP $h_{x}$. The $y_l$ is computed as $y_l=h_y(v_j-v_i,v_j,v_i)$, with an MLP $h_y$. Optionally, we utilize the obstacle encoding to update the $x$ and $y$. With Equation \ref{obsenc}, the $x$ and $y$ will merge the information from obstacles through multiple attention blocks, which is set as 3 in our experiments.

As mentioned in the main part, the GNN explorer will update $x$ and $y$ over multiple loops. During training, we iterate $x$ and $y$ over a random number of loops between 1 and 10. Intuitively, taking random loops encourages the GNN to converge faster, which also helps propagating the gradient. During evaluation, the GNN explorer will output $x$ and $y$ after 10 loops. For loops larger than 10, significant improvement on performance is not perceived. Finally, with an MLP $f_\eta$, the GNN explorer will output $\eta=f_{\eta}(y)$, which will be used as the priority to explore corresponding edges.

\paragraph{Special features in smoother architecture.} The GNN smoother embeds vertices with $x=h_x(v)$, with an MLP $h_{x}$. The $y_l$ is computed as $y_l=h_y(v_j-v_i,v_j,v_i)$, with an MLP $h_y$. Each time $x$ and $y$ are updated, the GNN smoother will output a new smoother path $\pi'=\{(u_i, u_i')\}_{i\in[0,k]}$ , where $u_i=f_u(x_i), \forall v_i \in \pi$, given an MLP $f_u$. The $u_0$ and $u'_{k}$ are manually replaced by $v_s$ and $v_g$, to satisfy the path constraint. We assume the new smoother path has the same number of nodes as the original path. Since the GNN smoother could gain novel local geometric information with the changed configuration of the new path, we dynamically update $G=\langle V,E\rangle$, via (i) replacing those nodes labeled as path nodes in $V$ by the nodes on new path, (ii) replacing $E$ by generating a k-NN graph on the updated $V$. With the updated graph $G$, we repeat the above operation, and subsequently get another new path, which forms a loop. By updating the graph and the new path iteratively and dynamically, the path is potentially improved to be shorter at each round by perceiving the changing local neighbors. During training, the GNN smoother outputs $\pi'$, after a random number of loops, which is between 1 and 10. During evaluation, the GNN smoother will output $\pi'$ after only one loop, but will be called 5 times in total for each smoothing tasks.

\section{Environments and Datasets}

\begin{figure*}[!htb]  
\centering
\includegraphics[width=0.8\textwidth]{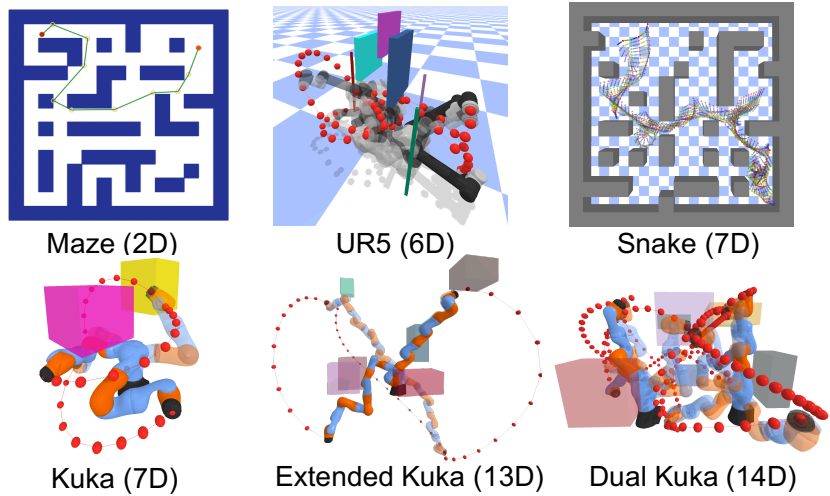}
\caption{Demonstrations of all our environments.}

\end{figure*}



We conduct the experiment on 6 different environments, which are described in details as follows, :

\paragraph{Maze} The maze contains a 2D point robot. The datasets for training set and the test set for "Easy2D" is at \url{https://github.com/NeurEXT/NEXT-learning-to-plan/tree/master/algorithm}~\cite{NEXT}. To generate the "Hard2D", we utilize the script provided by \url{https://github.com/RLAgent/gated-path-planning-networks}~\cite{GPPN}. The Hard mazes are generated by controlling the obstacle density not less than 46\%, and the distance from start to goal not less than 1.

\paragraph{UR5} The UR5 contains a UR5 robot arm ~\cite{Transporter}, which has 6 degrees of freedom. There are two sets of boxes, poles and pads, which are set to generate in two different size range. The poles and pads are randomly generated in the workspace for each problem.

\paragraph{Snake} The Snake environment contains a snake robot with 5 sticks, with another 2 degrees for the end position, which means 7D in total. The mazes are the same set of 2D mazes from NEXT.

\paragraph{Kuka, Extended Kuka} The Kuka environment contains a 7DoF Kuka arm with fixed base position. The extended Kuka environment contains an extended 13DoF kuka arm. The boxes are randomly generated in the workspace for each problem.

\paragraph{Dual Kuka} The environment contains two 7DoF KUKA arms, with 14DoF in total. Each arm need to reach the goal configuration, while required to not only avoid collision with the obstacles but also the other arm.

All the environments except the mazes are all implemented by PyBullet~\cite{pybullet} with the MIT license. The URDF files are contained in our supplementary codes.


\section{Tables for Overall Performance}

Here we list the overall performances of all the methods on all the environments, including the averaged value with the standard deviation.

\begin{table}[!htb]
\caption{Success rate. Our algorithm benefits from the probabilistic complete property from the RGG, which samples uniformly from free space,}
~\\
\centerline{\resizebox{\columnwidth}{!}{
	\begin{tabular}{c|c|c|c|c|c|c|c}
	\hline
		 & Easy2D & Hard2D & UR5 & Snake & Kuka7D & 13D & 14D \\
		\hline
		GNN & \textbf{1.00±0.00} & \textbf{1.00±0.00} & 0.96±0.00 & \textbf{1.00±0.00} & 0.99±0.00 & 0.99±0.00 & 0.99±0.00 \\
		\hline
		GNN + Smoother & \textbf{1.00±0.00} & \textbf{1.00±0.00} & 0.96±0.00 & \textbf{1.00±0.00} & 0.99±0.00 & 0.99±0.00 & 0.99±0.00 \\
		\hline
		GNN w/o OE & \textbf{1.00±0.00} & \textbf{1.00±0.00} & 0.96±0.00 & \textbf{1.00±0.00} & 0.99±0.00 & 0.99±0.00 & 0.99±0.00 \\
		\hline
		GNN w/o OE + Smoother & \textbf{1.00±0.00} & \textbf{1.00±0.00} & 0.96±0.00 & \textbf{1.00±0.00} & 0.99±0.00 & 0.99±0.00 & 0.99±0.00 \\
		\hline
		BIT* & \textbf{1.00±0.00} & \textbf{1.00±0.00} & \textbf{0.99±0.00} & \textbf{1.00±0.00} & \textbf{1.00±0.00} & \textbf{1.00±0.00} & \textbf{1.00±0.00} \\
		\hline
		NEXT & 0.99±0.00 & 0.97±0.00 & 0.37±0.00 & 0.72±0.01 & 0.88±0.01 & 0.61±0.01 & 0.67±0.00 \\
		\hline
		RRT* & 0.87±0.00 & 0.54±0.01 & 0.39±0.00 & 0.69±0.00 & 0.83±0.00 & 0.67±0.01 & 0.70±0.00 \\
		\hline
		LazySP & \textbf{1.00±0.00} & \textbf{1.00±0.00} & \textbf{0.99±0.00} & \textbf{1.00±0.00} & 0.99±0.00 & 0.99±0.00 & 0.99±0.00 \\
        \hline		
	\end{tabular}
}}
\label{table:success_rate}
\end{table}

\begin{table}[!htb]
\caption{Collision check. GNN performs the best in most high dimensional problems.}
~\\
\centerline{
		\resizebox{\columnwidth}{!}{\begin{tabular}{c|c|c|c|c|c|c|c}
		\hline
			 & Easy2D & Hard2D & UR5 & Snake & Kuka7D & 13D & 14D \\
			\hline
			GNN & 336.25±3.71 & 715.65±6.76 & \textbf{2474.03±40.35} & 1602.16±22.66 & \textbf{350.52±8.29} & \textbf{521.70±44.58} & \textbf{486.95±12.99} \\
			\hline
			GNN + Smoother & 496.79±4.68 & 1029.72±8.33 & 5182.02±191.40 & 2813.75±15.01 & 477.32±9.06 & 830.95±49.06 & 791.78±14.27 \\
			\hline
			GNN w/o OE & 332.30±4.00 & \textbf{703.72±5.78} & 2556.63±49.91 & 1605.73±24.00 & 353.89±7.47 & 588.65±51.04 & 547.16±37.61 \\
			\hline
			GNN w/o OE + Smoother & 565.38±6.37 & 1126.02±9.91 & 3715.40±132.41 & 2757.88±62.64 & 466.06±6.70 & 820.70±55.97 & 789.25±36.54 \\
			\hline
			BIT* & 478.88±10.95 & 1253.56±15.38 & 4055.73±286.93 & 1612.22±78.85 & 1951.81±424.82 & 1175.42±287.68 & 1276.95±230.88 \\
			\hline
			NEXT & \textbf{270.23±13.92} & 1206.09±18.62 & 6461.13±14.31 & 4788.84±20.60 & 2488.49±33.76 & 4958.80±99.51 & 4559.99±21.92 \\
			\hline
			RRT* & 1785.46±27.93 & 4080.07±32.69 & 3135.36±4.03 & 3352.45±15.68 & 1698.04±28.34 & 3004.45±55.36 & 2796.99±13.89 \\
			\hline
			LazySP & 351.80±2.47 & 801.21±6.74 & 2742.12±113.08 & \textbf{1595.74±48.15} & 369.36±19.42 & 546.64±29.40 & 604.64±38.84 \\
			\hline			
		\end{tabular}
}}
	
	\label{table:collision_check}
\end{table}

\begin{table}[!htb]
\caption{Path cost. With the GNN smoother, our path cost is the lowest from UR5 to 14D.}
~\\
\centerline{\resizebox{\columnwidth}{!}{
		\begin{tabular}{c|c|c|c|c|c|c|c}
		\hline
			 & Easy2D & Hard2D & UR5 & Snake & Kuka7D & 13D & 14D \\
			\hline
			GNN & 1.34±0.01 & 2.39±0.02 & 4.54±0.17 & 4.33±0.05 & 9.15±0.11 & 15.91±0.15 & 15.26±0.21 \\
			\hline
			GNN + Smoother & 1.18±0.01 & 2.05±0.01 & \textbf{4.12±0.12} & \textbf{3.91±0.01} & \textbf{6.14±0.02} & \textbf{8.98±0.06} & \textbf{8.86±0.08} \\
			\hline
			GNN w/o OE & 1.36±0.01 & 2.41±0.03 & 4.50±0.15 & 4.35±0.03 & 9.00±0.04 & 15.52±0.06 & 15.34±0.14 \\
			\hline
			GNN w/o OE + Smoother & 1.46±0.01 & 2.45±0.03 & 4.45±0.15 & 4.43±0.02 & 8.18±0.06 & 12.91±0.13 & 12.59±0.03 \\
			\hline
			BIT* & 1.11±0.00 & 2.00±0.02 & 4.33±0.09 & 3.95±0.02 & 6.57±0.04 & 9.41±0.07 & 9.54±0.04 \\
			\hline
			NEXT & \textbf{1.02±0.00} & \textbf{1.71±0.01} & 4.62±0.05 & 5.45±0.04 & 7.74±0.06 & 10.17±0.07 & 10.66±0.12 \\
			\hline
			RRT* & 1.14±0.01 & 1.79±0.01 & 4.66±0.03 & 4.69±0.05 & 6.95±0.02 & 9.81±0.04 & 10.52±0.03 \\
			\hline
			LazySP & 1.20±0.01 & 2.11±0.01 & 4.30±0.06 & 4.18±0.03 & 8.16±0.05 & 13.51±0.06 & 13.54±0.06 \\
			\hline
		\end{tabular}
}}
	\label{table:path_cost}
\end{table}

\begin{table}[!htb]
\caption{Total running time. GNN requires low time cost due to its optimization on collision checks.}
~\\
\centerline{\resizebox{\columnwidth}{!}{
		\begin{tabular}{c|c|c|c|c|c|c|c}
		\hline
			 & Easy2D & Hard2D & UR5 & Snake & Kuka7D & 13D & 14D \\
			\hline
			GNN & \textbf{35.07±0.78} & 80.57±1.16 & \textbf{290.41±3.20} & 218.83±3.41 & \textbf{56.77±1.89} & \textbf{102.67±10.48} & \textbf{84.25±3.16} \\
			\hline
			GNN + Smoother & 48.71±0.79 & 99.40±1.18 & 481.31±13.31 & 312.30±2.64 & 72.32±1.93 & 143.71±11.00 & 119.68±2.95 \\
			\hline
			GNN w/o OE & 35.43±0.66 & \textbf{80.19±1.04} & 298.18±4.25 & 221.26±2.29 & 57.62±1.87 & 116.69±11.75 & 95.19±8.36 \\
			\hline
			GNN w/o OE + Smoother & 51.59±0.76 & 101.40±1.17 & 386.30±9.97 & 310.42±4.39 & 72.39±1.85 & 149.86±12.19 & 125.09±8.25 \\
			\hline
			BIT* & 47.69±3.43 & 146.55±2.88 & 387.44±27.34 & \textbf{183.84±6.69} & 199.19±42.31 & 183.54±40.22 & 167.81±21.43 \\
			\hline
			NEXT & 166.46±4.81 & 499.65±4.40 & 7150.17±690.44 & 4355.78±28.44 & 1837.28±40.04 & 4750.26±88.29 & 4450.67±278.09 \\
			\hline
			RRT* & 77.20±1.03 & 166.06±1.35 & 396.86±0.45 & 425.87±2.46 & 166.98±2.43 & 465.80±8.58 & 392.57±2.74 \\
			\hline
    		LazySP & 83.65±1.97 & 287.30±11.93 & 905.22±97.21 & 236.79±31.76 & 339.97±61.63 & 301.63±67.92 & 465.30±93.14 \\		
    		\hline			
		\end{tabular}
}}
	\label{table:total_time}
\end{table}

\newpage

\section{Breakdown of the Total Planning Time}
\begin{figure*}[h!]  
\centering
\includegraphics[width=\textwidth]{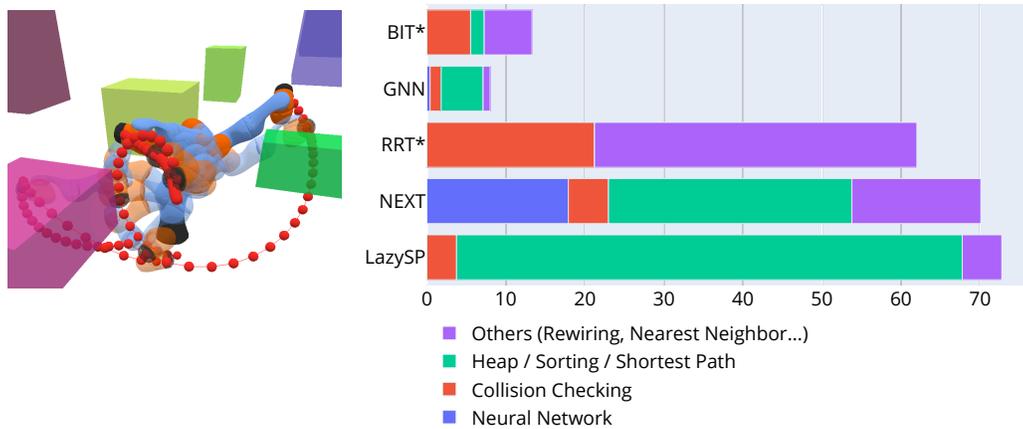}
\caption{On the left environment, we breakdown the total planning time into various operations. The time taken by each operation is shown on the right chart.}
\label{figure:decompose}
\end{figure*}

As shown in Figure~\ref{figure:decompose}, we breakdown the total planning time into various operations on an example environment. We found that our method takes most of the time sorting the priority on the frontier. On contrary, BIT* and RRT* take relatively large amount of time checking for collisions. NEXT needs to recalculate the exploration bonus, and sort the candidates to explore, which takes prohibitive computation. LazySP searches for global shortest path every time an edge is in collision, which makes the search on path become the bottleneck.

\section{Ablation Study}

\subsection{Varying Training Set Size for Explorer}

\begin{figure}[!hbt]  
\begin{center}
\centerline{\includegraphics[width=\textwidth]{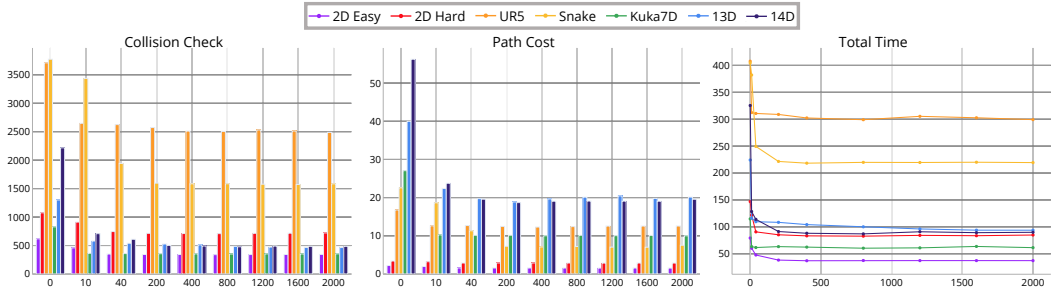}}
\caption{Ablation study on the different training set size. All the performances become stable at relatively few training set size (around 40 problems, 2\% of the original training set).}
\label{training set}
\vspace{-0.7cm}
\end{center}
\end{figure}

We conduct further experiments to analyze the effect of the training set size. We train the GNN path explorer with the 0 (0\%), 10 (0.5\%), 40 (2\%), 200 (10\%), 400 (20\%), 800 (40\%), 1200 (60\%), 1600 (80\%), 2000 (100\%) problems in this new training set. The performance of collision checks, path costs, and total time on the testing problems are demonstrated in Figure \ref{training set}.

As shown, the overall performance of the GNN explorer is robust, even with 2\% problems of the original training set. There are two reasons here: (i) The GNN models we propose are relatively lightweight in terms of the parameter numbers, which means that it is suffice to train it with small amount of data. (ii) Our GNN model does not depend on the global feature of the whole graph, as it only aggregates the information from local neighborhoods. Though each problem yields a different graph in terms of global characteristic, they can share similar local geometric patterns, which is beneficial for the efficiency of learning GNN models.

\subsection{Feature Choices}

\begin{figure*}[ht]  
\centering
\centering
\hbox{\hspace{9em}\includegraphics[width=0.7\textwidth]{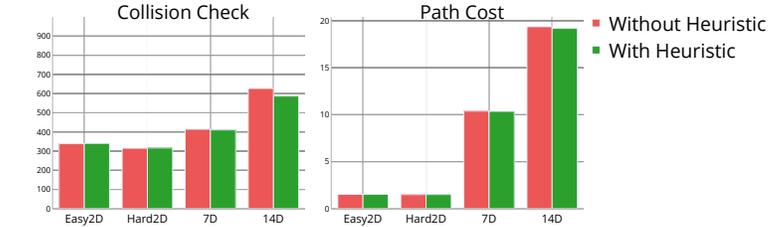}}
\caption{Comparison of performance of GNN explorers with vertex embedding $x$ as $h_x(v, v_g)$ and $h_x(v, v_g, (v-v_g)^2, v-v_g)$ respectively. Results show that the GNN explorer with heuristics perform slightly better for high-dimensional problems.}
\label{feature}
\vspace{-0.5cm}
\end{figure*}

In this experiment, we replace the vertex embedding $x=h_x(v, v_g, (v-v_g)^2, v-v_g)$ by $x=h_x(v, v_g)$, which removes the L2 distance heuristic on features. We retrain the new GNN with the same training set, and compare to the original architecture on 4 environments. As shown in Figure~\ref{feature}, the performances of two GNNs are close to each other from 2D to 7D environments (0.5\%, 0.6\%, 1.0\% in terms of collision checking), and the original GNN is slightly better on 14D environment (6.8\% in terms of collision checking, 0.9\% in terms of path cost). The L2 distance heuristic is helpful in high dimensions, but does not have much effect, due to the complex geometry of the C-space.

\subsection{GNN Smoother Versus Oracle Smoother}

\begin{figure*}[ht]  
\centering
\centering
\includegraphics[width=0.9\textwidth]{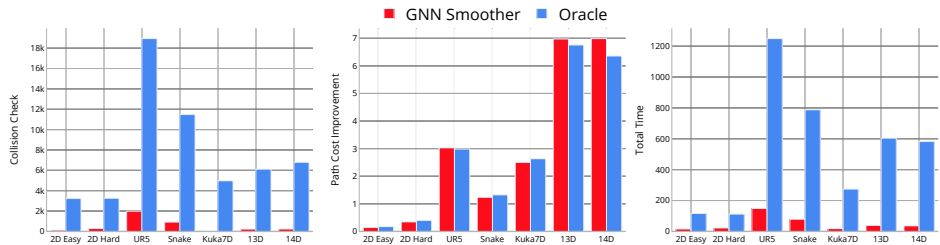}
\caption{Comparison of performance between our GNN smoother and the oracle to trained on. Our GNN smoother learns to smooth the path with comparable improvement as the oracle, and also requires fewer collision checking steps and less total time.}
\label{path smoother}
\end{figure*}

In this experiment, we replace the learned GNN smoother by the oracle smoother, which is the expert that GNN smoother imitates. We compare these two smoothers, given the same path explored by the GNN explorer on the test problems. As shown in Figure \ref{path smoother}, our smoother requires much fewer collision checks and time, while maintaining comparable improvement on the explored path, which is contributed by the incremental way of smoothing, and the generalizablity of the GNN. More specifically, the GNN smoother requires 3.9\%, 8.8\%, 10.5\%, 8.0\%, 1.9\%, 3.7\%, 3.5\% as many collision checks as the oracle on each environment, while maintaining 80.7\%, 86.9\%, 101.5\%, 93.2\%, 94.9\%, 103.2\%, 109.9\% as much improvement as the oracle for each environment.


\subsection{Varying the $k$ in k-NN}

\begin{figure}[!hbt]  
\begin{center}
\centerline{\includegraphics[width=\textwidth]{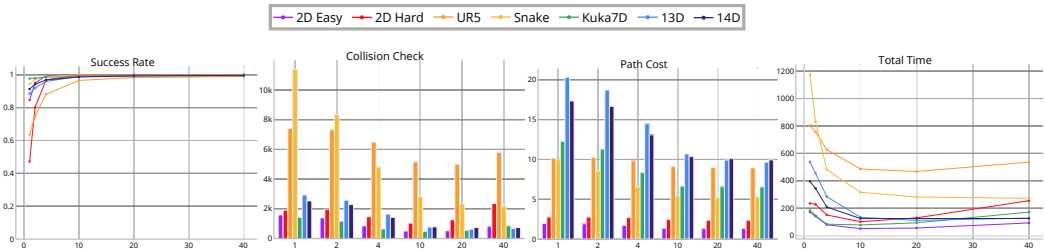}}
\caption{Ablation study on different $k_0$ for k-NN. Performances are best at $k_0 \in [10, 20]$.}
\label{k number}
\vspace{-0.7cm}
\end{center}
\end{figure}

As suggested in \citet{k-NN}, we set the $k$ in the k-NN graph as proportional to the logarithm of the number of vertices, which is formulated as $\lceil k_0\cdot \frac{\log{|V_{f}|}}{\log{100}}\rceil$. Here we test different $k_0$ for the k-NN, choosing among $\{1,2,4,10,20,40\}$, as demonstrated in Figure \ref{k number}.

It is not surprising for the success rate to increase when $k$ increases, since there are more edges in the graph, which increases the possibility to find a feasible path. The path cost also decreases with increasing $k$, since on average there might be fewer segments on a path. The collision checks and total running time first decrease then increase, since larger $k$ brings higher possibility to find a feasible path with fewer intermediate vertices, but also brings more edges to check on the frontier.


\subsection{Varying the Batch Size}

In this experiment, we inspect the effect of the batch size. While constraining the maximum sampling number from free space to be 1000, we set the batch sampling size among $\{50,100,200,250,500,1000\}$. We see that the success rate drops when the batch size increases, since the GNN explorer is given fewer opportunities to fail and re-sample. The collision checks grows with the batch size, because the graphs would contain more edges with larger batch size on average. The path cost is lower with larger batches, similar to the effect of higher $k$, due to higher possibility to find a feasible path with fewer intermediate vertices. The total time raises when the batch size goes larger, because larger batches brings denser graphs, which enlarges both CPU and GPU costs. 

\begin{figure}[!hbt]  
\begin{center}
\centerline{\includegraphics[width=\textwidth]{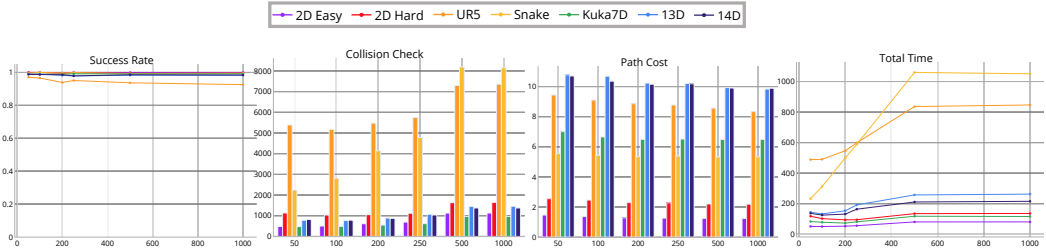}}
\caption{Ablation study on different batch size for batch sampling. Larger batches tend to yield lower success rate and path costs, while requiring more collision checks and running time.}
\label{batch size}
\vspace{-0.7cm}
\end{center}
\end{figure}

\section{Hyperparameters}

The hyperparameters that we use are listed in the following table. 

\begin{center}
 \begin{tabular}{c | c} 
 \hline
\textbf{Hyperparameters} & \textbf{Values} \\
 \hline
 Maximum sampling number & 1000 \\
  \hline
 $k$ for k-NN  & $\lceil 10\cdot \frac{\log{|V_{f}|}}{\log{100}}\rceil$ \\ 
 \hline  
 GNN batch size & 100 \\ 
 \hline
BIT* batch size  & 100 \\
 \hline
 RRT*/NEXT step size on 2D & 5e-2  \\
 \hline
 RRT*/NEXT step size on 7D/13D/14D & 5e-1 \\
 \hline
 Training epoch & 20\\
 \hline
 Training batch size & 8\\
 \hline
 Learning rate & 1e-3\\
 \hline
 Random seeds & 1234, 2341, 3412, 4123\\
 \hline
\end{tabular}
\end{center}

\section{Algorithms}

\label{section:psd}



\begin{algorithm}[!htb]
   \caption{GNNExplorer}
   \label{alg:explore}
\begin{algorithmic}
   \STATE {\bfseries Input:} obstacles $O$, start $v_s$, goal $v_g$, batch size $n$, node limit $T_{max}$
   \STATE Sample $n$ nodes from $C_{free}$ to $V_f$
   \STATE Sample $n$ nodes from $C_{obs}$ to $V_c$
   \STATE Initialize $G = \{V:\{v_s, v_g\}\cup V_f \cup V_c, E:\text{k-NN}(V_f)\cup \text{k-NN}(V)\}$
   \STATE Initialize $i=0, E_0=\text{k-NN}(V_f), V_{\mathcal{T}_{0}}=\{v_s\}, E_{\mathcal{T}_{0}}=\emptyset$
   \STATE $\eta= \mathcal{N}_E (V, E, O)$
   \REPEAT
   \STATE select $e_i$ with $\eta$
   \STATE $E_i \leftarrow E_i \setminus \{e_i\}$
   \IF{$e_i:(v_i, v_i')\subseteq C_{free}$} 
        \STATE $V_{\mathcal{T}_{i+1}} \leftarrow V_{\mathcal{T}_i} \cup\{v_i'\}$
        \STATE $E_{\mathcal{T}_{i+1}} \leftarrow E_{\mathcal{T}_i} \cup\{e_i\}$
        \STATE $E_{i+1} \leftarrow E_{i}$
        \STATE $E_{f}(\mathcal{T}_{i+1})\leftarrow\{e_j:(v_j, v_j')\in E_{i+1} \mid v_j\in V_{\mathcal{T}_{i+1}}, v_j'\not\in V_{\mathcal{T}_{i+1}}\}$
        \STATE $i\leftarrow i+1$
        \IF{$||v_i'-v_g||_2^2 \leq \delta$}
            \STATE $\pi \leftarrow$ path from $v_s$ to $v_g$ on tree $\mathcal{T}_{i}$
            \RETURN $\pi$
        \ENDIF
   \ENDIF
   \IF{$ E_i \cap E_{f}(\mathcal{T}_i)==\emptyset$}
       \STATE Sample $n$ nodes from $C_{free}$, add to $V_f$
       \STATE Sample $n$ nodes from $C_{obs}$, add to $V_c$
        \STATE $V \leftarrow \{v_s, v_g\}\cup V_f \cup V_c$
        \STATE $E_i \leftarrow \text{k-NN}(V_f) \setminus (E\setminus E_i)$
        \STATE $E \leftarrow \text{k-NN}(V_f)\cup \text{k-NN}(V)$
        \STATE $\eta = \mathcal{N}_E ( V, E, O)$
   \ENDIF
   \UNTIL{$|V_f| > T_{max}$} 
   \RETURN $\emptyset$
\end{algorithmic}
\end{algorithm}

\begin{algorithm}[!htb]
   \caption{GNNSmoother}
   \label{alg:smoother}
\begin{algorithmic}
  \STATE {\bfseries Input:} step size $\epsilon$, stop difference $\delta$, outer loop $L$, inner loop $K$
   \STATE {\bfseries Input:} explored path $\pi:(v_i,v_i')_{i\in[0,k]}$, free samples $V_f$, collided samples $V_c$  
   \FOR{$n \in \{1\dots L\}$}
   \STATE $G \leftarrow \{V:\{V_\pi, V_f, V_c\}, E:\text{k-NN}(V_\pi, V)\cup E_\pi\}$
   \STATE $\pi':(u_i,u_i')_{i\in[0,k]} = \mathcal{N}_S(V, E)$
   \FOR{$m \in \{1\dots K\}$}
   \STATE  $d \leftarrow 0$
   \FOR{$u_i\in V_{\pi'}, i \in [1, k]$}
   \STATE $w_i \leftarrow$ steer $v_i$ toward $u_i$ within step $\epsilon$
   \IF{$e_{i-1}: (v_{i-1}, w_i) \subseteq C_{free}$}
   \STATE Replace $v_i\in \pi$ with $w_i$
   \STATE  $d \leftarrow d + ||w_i-u_i||^2_2$
   \ENDIF
    \ENDFOR
\IF{$d \leq \delta$}
            \STATE \textbf{break}
            \ENDIF
    \ENDFOR
   \ENDFOR
   \RETURN $\pi$
\end{algorithmic}
\end{algorithm}


\begin{algorithm}[hbt!]
   \caption{RandomSmoother}
   \label{alg:randomSmoother}
\begin{algorithmic}
   \STATE {\bfseries Input:} path $\pi:(v_i,v_i')_{i\in[0,k]}$, perturbation range $\epsilon$, iteration $L_R$
    \FOR{$n \in \{1\dots L_R\}$}
        \STATE pick a random node $v_i \in \pi, 1\leq i \leq k$
        \STATE $u_i \leftarrow v_i + \text{random}(-\epsilon, \epsilon)$
        \IF{$(v_{i-1}, u_i), (u_i, v_{i+1}) \subseteq C_{free}$ and $Cost[(v_{i-1}, u_i), (u_i, v_{i+1})]< Cost[(v_{i-1}, v_i), (v_i, v_{i+1})]$}
        \STATE replace $v_i$ with $u_i$
        \ENDIF
    \ENDFOR
    \RETURN $\pi$
\end{algorithmic}
\end{algorithm}

\begin{algorithm}[hbt!]
   \caption{SegmentSmoother}
   \label{alg:segmentSmoother}
\begin{algorithmic}
   \STATE {\bfseries Input:} perturbed path $\pi:(v_i,v_i')_{i\in[0,k]}$ from Algorithm \ref{alg:randomSmoother}
   \STATE $critical= [v_0, v_k']$
    \FOR{$i \in [1,k]$}
        \IF{$(v_{i-1}, v_i') \not \subseteq C_{free}$}
        \STATE append $v_i$ to $critical$
        \ENDIF
    \ENDFOR
    \STATE $\pi_M = \emptyset$
    \FOR{adjacent pair $v_i, v_j\in critical$}
        \STATE $V \leftarrow \{v_p\mid v_p\in \pi, i\leq p \leq j\}$
        \STATE $E \leftarrow \{(v_a, v_b)\mid v_a, v_b \in V, (v_a, v_b)\subseteq C_{free}\}$
        \STATE $\pi_{ij}\leftarrow$ the shortest path from $v_i$ to $v_j$ via Dijkstra$(V, E)$
        \STATE $\pi_M \leftarrow  \pi_M \cup \pi_{ij}$
    \ENDFOR
    \RETURN $\pi_M$
\end{algorithmic}
\end{algorithm}

The smoothing oracle that we use is similar to the approach of gradient-informed path smoothing proposed by ~\citet{GradSmooth}. Since the gradient in the configuration space is complex, we replace the gradient smoother by a random perturbation smoother. The oracle smoother jointly calls the random perturbation smoother and a segment smoother over multiple iterations. These two smoothers are described in Algorithm \ref{alg:randomSmoother} and \ref{alg:segmentSmoother}.

\FloatBarrier
\bibliography{main}
\bibliographystyle{abbrvnat}